\documentclass[lettersize,journal]{IEEEtran}
\usepackage[hyphens]{url}  
\usepackage{graphicx} 
\usepackage[numbers,sort&compress]{natbib}
\usepackage{caption} 
\usepackage{booktabs}
\usepackage{amsmath}
\usepackage{amsfonts}
\usepackage{bm}
\usepackage[table]{xcolor}
\usepackage{multirow}
\usepackage{booktabs}

\newcommand{\method}{CSTF}
\newcommand{\tmtwo}{T2M}
\newcommand{\era}{ERA5}
\newcommand{\eraland}{ERA5-Land}

\definecolor{bestgreen}{HTML}{B9FBC0}
\newcommand{\best}[1]{\cellcolor{bestgreen}\textbf{#1}}
\newcommand{\second}[1]{\underline{#1}}

\usepackage{amsmath}
\usepackage{amssymb}
\usepackage{mathtools}
\usepackage{bm}

\DeclareMathOperator{\MLP}{MLP}
\DeclareMathOperator{\SinCos}{SinCos}

\newcommand{\bX}{\mathbf{X}}
\newcommand{\bY}{\mathbf{Y}}
\newcommand{\bZ}{\mathbf{Z}}
\newcommand{\bP}{\mathbf{P}}
\newcommand{\bR}{\mathbf{R}}

\newcommand{\bp}{\mathbf{p}}
\newcommand{\br}{\mathbf{r}}

\newcommand{\hY}{\widehat{\mathbf{Y}}}
\newcommand{\bphi}{\boldsymbol{\phi}}

\begin{document}

\title{Learning Continuous Regional Temperature Fields with Lead-Time and Resolution Queries}

\author{Chunlei~Shi, Jiong~Wang, Yi-Lin~Wei, Junming~Hou, Jinjin~Liu, Yecheng~Zhang, \\and Dan~Niu,~\IEEEmembership{Member,~IEEE}

\thanks{This work was supported by the  Heavy Rainfall Research Foundation of China (No. BYKJ2025M14), China Meteorological Administration Xiong\textrm{'}an Atmospheric Boundary Layer Key Laboratory (No. 2025LABL-B12), and by the National Natural Science Foundation of China (62374031, 62331009), and by NSFC-Jiangsu Province (BK20240173). (\textit{Corresponding author: Chunlei Shi and Dan Niu.})}
\thanks{Chunlei Shi, Jinjin Liu and Dan Niu are with the Department of Automation, Southeast University, Nanjing 210096, China (e-mail: 230238514@seu.edu.cn, danniu1@163.com).}
\thanks{
Jiong Wang is with the Department of Information Science and Technology, Fudan University, Shanghai 200433, China.
Yi-Lin Wei is with the School of Computer Science and Engineering, Sun Yat-sen University, Guangzhou 510006, China. 
Junming Hou is with the State Key Laboratory of Millimeter Waves, School of Information Science and Engineering, Southeast University, Nanjing 210096,China. 
Yecheng Zhang is with the Department of Architecture, Tsinghua University, Beijing 100084, China.}}

\markboth{Journal of \LaTeX\ Class Files,~Vol.~14, No.~8, August~2021}%
{Shell \MakeLowercase{\textit{et al.}}: A Sample Article Using IEEEtran.cls for IEEE Journals}


\maketitle
\begin{abstract}
Accurate regional near-surface temperature forecasting is fundamental to short-range weather services and downstream risk assessment.
Existing deep learning-based regional forecasters commonly produce a fixed set of future frames on a prescribed grid, limiting their use when forecast products must be evaluated at query-dependent lead times or display resolutions.
To overcome these fixed-output constraints, we formulate regional \tmtwo{} forecasting as query-conditioned continuous spatiotemporal temperature field evaluation and propose the \textbf{C}ontinuous \textbf{S}patiotemporal \textbf{T}emperature \textbf{F}orecaster (\textbf{\method}), a neural field that turns forecast lead time and output resolution into explicit queries when evaluating 2-m temperature (T2M).
Specifically, \method{} first encodes multi-variable \era{} histories into latent meteorological states and then decodes \tmtwo{} as a coordinate-based field.
Accordingly, spatial location, forecast lead time, and output resolution are introduced as queries, enabling standard hourly forecasts, intermediate lead-time diagnostics, and resolution-controllable outputs within a unified field-evaluation framework.
Furthermore, to maintain coherence across flexible field queries, we design spatial-gradient, temporal-difference, and scale-consistency objectives that regularize regional thermal structures, lead-wise evolution, and cross-resolution agreement.
Experiments on the Southeast China 0--6 h \eraland{} benchmark demonstrate that \method{} achieves the best aggregate deterministic skill, including a 17.0\% reduction in Bias, with global-scope diagnostics further illustrating flexible lead-time and resolution-controllable inference.
\end{abstract}


\section{Introduction}

Accurate regional near-surface temperature forecasting supports weather services and decisions in electricity demand management, agricultural planning, public health warning, and urban heat-risk assessment \citep{mukkavilli2023ai,zhou2026reconstruction}.
Near-surface temperature evolves continuously under advection, radiation, cloud cover, land--sea contrast, and surface heterogeneity, so local gradients can be as important as domain-mean error for regional use \citep{arjdal2026compound,chajaei2024machine,liu2024mambads}.
Accordingly, practical forecast products are not limited to a fixed set of hourly maps on a single grid, but often involve intermediate horizons, coarse-resolution diagnostics, and denser-grid outputs for visualization or downstream coupling.

\begin{figure*}[t]
  \centering
  \includegraphics[width=\textwidth]{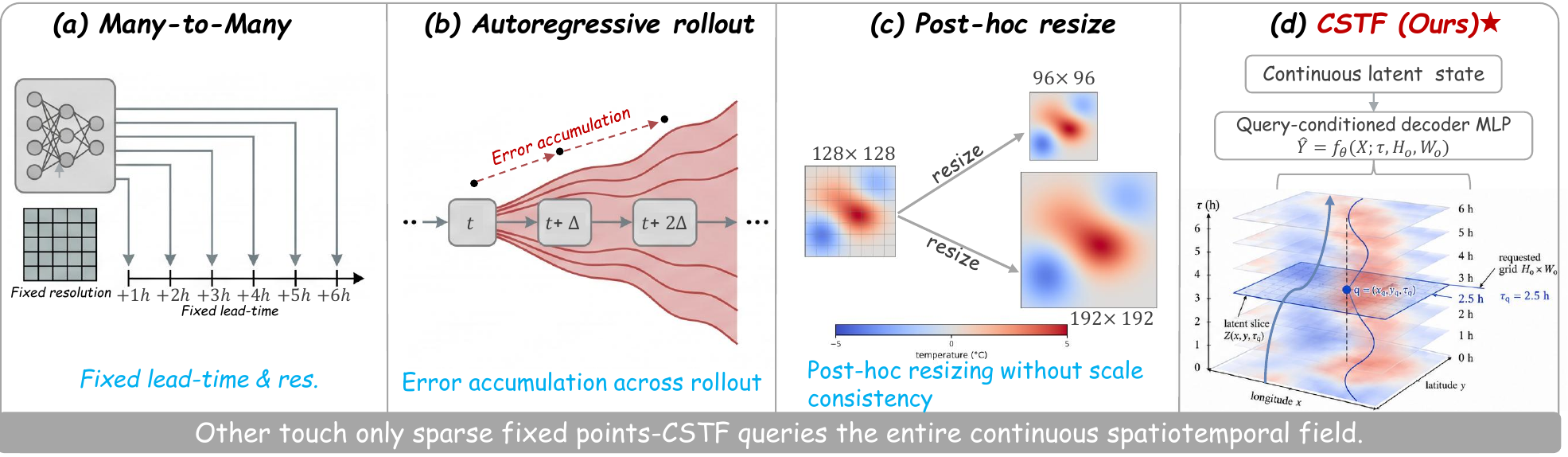}
  \caption{
  Comparison between conventional fixed-output and proposed query-conditioned temperature forecasting paradigms.
  (a) Fixed multi-head models predict predefined lead times and output resolutions.
  (b) Autoregressive rollout extends forecasts step by step but can accumulate errors across forecast steps.
  (c) Post-hoc resizing changes the display grid without enforcing consistency across resolutions.
  (d) Our \method{} formulates regional \tmtwo{} prediction as continuous field evaluation conditioned on spatial, lead-time, and resolution queries, enabling flexible forecast requests.
  }
  \label{fig:motivation}
\end{figure*}
Data-driven meteorological models have made rapid progress in learning structured atmospheric fields from large-scale gridded data \citep{bi2023accurate,lam2023learning,chen2023fengwu}.
Yet most regional neural forecasters still expose a discrete product interface, either emitting a fixed stack of future frames or advancing the state autoregressively at a predefined time step \citep{bi2023accurate,lam2023learning,chen2023fengwu}.
Such designs are natural for benchmark evaluation, but they make lead time and output resolution architectural choices rather than query variables.
A fixed multi-head or sequence-to-sequence model can return the frames it was built to produce, while a request such as $t+0.5$ h at $192\times192$ resolution lies outside its native output space (Fig.~\ref{fig:motivation}a).

Autoregressive rollout can extend the forecast horizon, but it does not by itself provide dense-time field evaluation and may accumulate errors as predictions are fed back into the model (Fig.~\ref{fig:motivation}b) \citep{lam2023learning,tian2025arrow}.
Likewise, post-hoc resizing or super-resolution can change the apparent grid after a forecast is generated, but it does not ensure that products sampled at different resolutions describe the same underlying temperature state (Fig.~\ref{fig:motivation}c) \citep{sun2024deep,hess2024fast,liu2024deriving,gao2023implicit}.
In all these cases, the available lead times and output grid are fixed by the training target or by the final prediction head.
Increasing the number of supervised frames or training separate models for different resolutions can enlarge the product catalogue, but the model still predicts among predefined outputs rather than evaluating a requested field.

This limitation motivates the central question of this work: can regional temperature forecasting be formulated as continuous spatiotemporal field evaluation \citep{verma2024climode,seong2026rainode}?
Instead of asking a network to emit a finite list of images, we seek a single field evaluator that maps the same encoded atmospheric state to \tmtwo{} values at requested spatial coordinates, lead times, and output grids.
Under this view, time and resolution are part of the model input, and cross-resolution consistency becomes a property that can be trained and diagnosed explicitly.

To this end, we propose the \textbf{C}ontinuous \textbf{S}patiotemporal \textbf{T}emperature \textbf{F}orecaster (\textbf{\method{}}), a query-conditioned neural field for regional 2-m temperature prediction (Fig.~\ref{fig:motivation}d).
\method{} first encodes multi-variable \era{} histories into a shared latent meteorological state, so forecast products at different times and grids are derived from the same atmospheric context.
A coordinate decoder then evaluates \tmtwo{} fields from explicit spatial-coordinate, lead-time, and resolution queries, enabling hourly forecasts, intermediate lead-time queries, and resolution-controllable outputs within one field-evaluation framework.
Unlike fixed prediction heads or separate scale-specific models, this design makes the requested lead time and grid part of the prediction process itself.
To make these flexible queries meteorologically coherent rather than merely interpolated products, we regularize regional thermal gradients, lead-wise temperature changes, and cross-resolution agreement during training.
We evaluate \method{} on a Southeast China 0--6 h \tmtwo{} forecasting benchmark and further use global-scope diagnostics to examine the same query interface beyond the regional test domain. Our contributions are summarized as follows:
\begin{itemize}
    \item We reformulate regional \tmtwo{} forecasting as query-conditioned continuous spatiotemporal field evaluation, where lead time and output resolution are treated as inputs rather than fixed output dimensions.
    \item We propose \method{}, a shared-latent-state and coordinate-decoding architecture that directly evaluates regional temperature fields at requested spatial locations, forecast lead times, and output grids.
    \item We develop coherence-preserving supervision and diagnostics for query-conditioned temperature fields, validating fixed-lead skill, continuous lead-time queryability, resolution control, and cross-resolution consistency.
\end{itemize}

\section{Related Work}
\begin{figure*}[t]
  \centering
  \includegraphics[width=\linewidth]{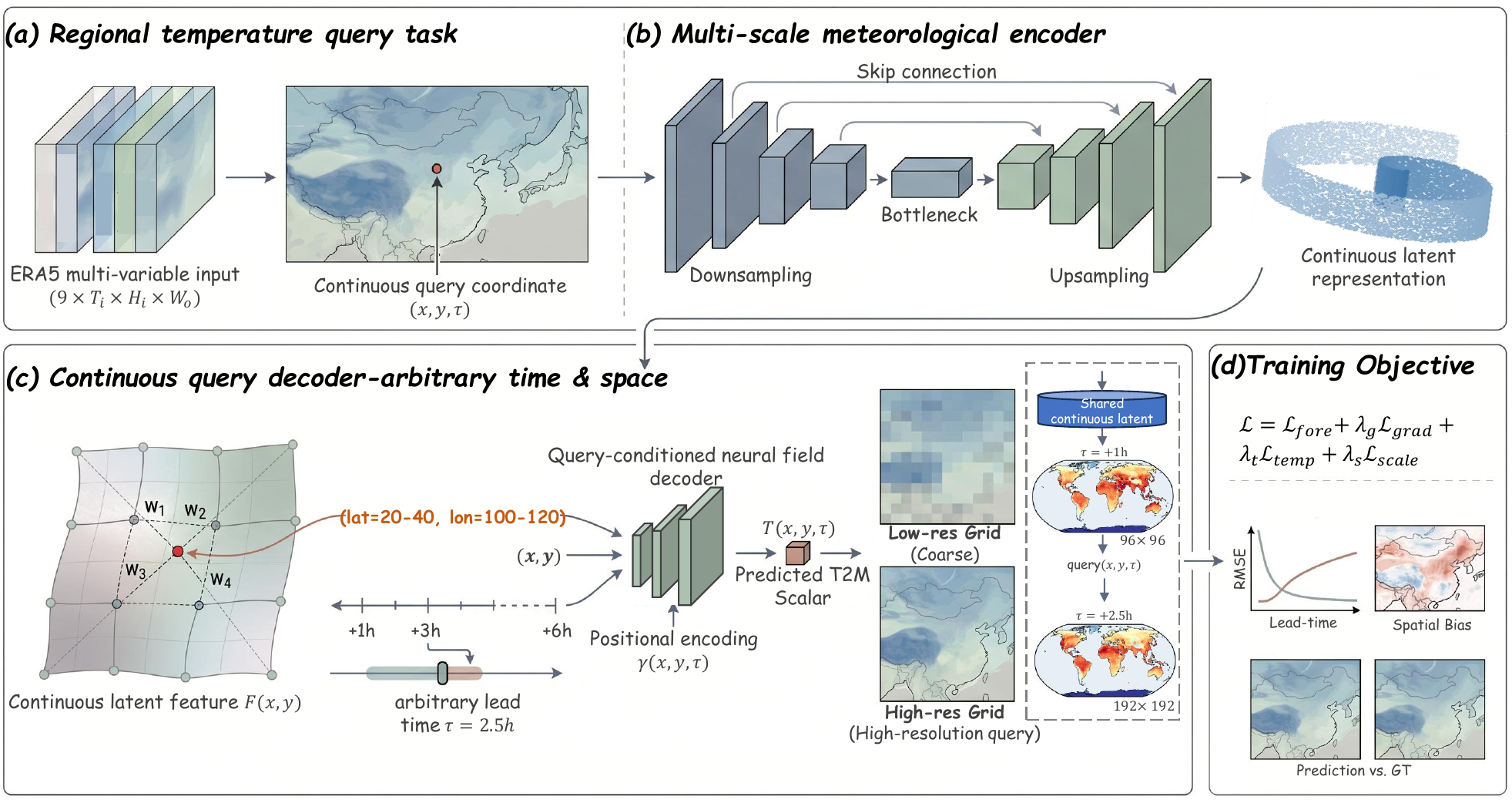}
  \caption{
  Overall architecture of the proposed \method{} framework.
  The model encodes \era{} multi-variable input history into a latent regional weather state within a unified query-conditioned forecasting pipeline.
  Conditioned on this latent state, a coordinate decoder predicts regional \tmtwo{} fields from lead-time, spatial-coordinate, and output-resolution queries.
  Training combines deterministic, gradient, temporal-difference, and scale-consistency objectives, so the same trained model supports continuous lead-time queries, resolution-controllable outputs, and scale-consistency evaluation.
  }
  \label{fig:framework}
  \vspace{-10pt}
\end{figure*}

\paragraph{Data-driven weather prediction.}
Machine learning is now widely used for weather and Earth-system prediction.
WeatherBench established a benchmark for data-driven global forecasting \citep{rasp2020weatherbench}, while FourCastNet, Pangu-Weather, GraphCast, and FengWu demonstrated skilful high-dimensional forecasts at low inference cost \citep{pathak2022fourcastnet,bi2023accurate,lam2023learning,chen2023fengwu}.
At shorter ranges and regional scales, EarthFormer, wavelet-driven radar retrieval, and attention-based convolutional models capture useful evolution patterns from multi-channel meteorological fields \citep{gao2022earthformer,shi2026wavec2r}.
However, most systems still use a fixed-output interface: input history is mapped to predefined variables, lead times, and grids.
Our work addresses this limitation for regional \tmtwo{} prediction by making lead time and output resolution query variables.

\paragraph{Regional temperature prediction.}
Regional near-surface temperature prediction is a structured short-range refinement problem governed by synoptic forcing, advection, radiation, cloud cover, land--sea contrast, topography, and surface heterogeneity \citep{arjdal2026compound,chajaei2024machine}.
\era{} provides dynamically coherent atmospheric context, whereas \eraland{} resolves regional thermal gradients relevant to local decisions \citep{hersbach2020era5,wang2026fusing}.
Convolutional encoder--decoders, transformer forecasters, and super-resolution models map coarse or multi-variable inputs to regional targets, but their lead times and grids are usually fixed during training \citep{gao2022earthformer,sun2024deep,hess2024fast,liu2024mambads}.
Post-hoc interpolation changes the display grid without ensuring cross-resolution agreement for the same temperature state \citep{liu2024deriving,gao2023implicit}; \method{} instead keeps supervised \eraland{} verification while adding inference-time control over lead time and resolution.

\paragraph{Continuous and coordinate-based modelling.}
Recent weather-AI models relax fixed temporal interfaces by treating evolution as continuous or adapting rollout intervals \citep{verma2024climode,liu2024mitigating,tian2025arrow}.
Temporal interpolation and temporal super-resolution have also been explored for radar precipitation and atmospheric fields \citep{tatsubori2022deep,demiray2023efficienttempnet,wang2025temdeep}.
Coordinate-based neural fields and operator-learning methods further represent signals or solution maps as functions evaluable at requested locations or discretizations \citep{li2020fourier,guibas2021adaptive}.
Together, these ideas motivate query-based meteorological prediction, yet temporal flexibility can leave the grid fixed, and coordinate flexibility alone does not ensure cross-resolution coherence.
\method{} couples lead-time, coordinate, and resolution conditioning for regional \tmtwo{} forecasting, using scale-consistency training and diagnostics to separate controllable inference from post-processing.

\subsection{Problem definition}

Given a regional domain $\Omega$ discretized on a base grid of size $H\times W$, each forecast initialization provides a multi-variable \era{} history
\begin{equation}
  \bX
  \in
  \mathbb{R}^{B\times T_{\mathrm{in}}\times C_{\mathrm{in}}\times H\times W},
\end{equation}
where $B$ is the batch size, $T_{\mathrm{in}}$ denotes the number of historical frames, and $C_{\mathrm{in}}=9$ corresponds to
\begin{equation}
  \{\mathrm{t2m}, \mathrm{d2m}, \mathrm{u10}, \mathrm{v10}, \mathrm{msl}, \mathrm{cp}, \mathrm{lsp}, \mathrm{tcc}, \mathrm{ssrd}\}.
\end{equation}
The supervised target is the future \eraland{} \tmtwo{} sequence $\bY\in\mathbb{R}^{B\times T_{\mathrm{out}}\times 1\times H\times W}$, defined at hourly lead times for training and standard evaluation.

The standard supervised formulation learns a fixed-output sequence mapping
\begin{equation}
  \bX \mapsto \bigl\{\bY_1,\ldots,\bY_{T_{\mathrm{out}}}\bigr\},
\end{equation}
where the available lead times and output grid are determined by the target sequence and prediction head.
Such a predictor can only return members of its predefined product set, and therefore cannot natively evaluate the temperature field at an intermediate lead time or an alternative grid.
To expose lead time and output resolution as query variables, we instead formulate regional near-surface temperature forecasting as query-conditioned spatiotemporal field evaluation:
\begin{equation}
  \hY_{\tau}^{(H_o,W_o)}
  =
  \mathcal{F}_{\theta}\!\left(\bX;\tau,H_o,W_o\right),
  \label{eq:field-evaluator}
\end{equation}
where $\tau$ denotes the requested lead time and $(H_o,W_o)$ denotes the requested output grid.
The objective is to learn a single function $\mathcal{F}_{\theta}$ that remains accurate at supervised integer leads while supporting diagnostic queries at non-integer lead times and multiple output resolutions.

\subsection{Overview}

To implement this field-evaluation formulation, merely appending lead-time and resolution inputs to a fixed-grid forecaster is insufficient.
If forecast products at different horizons or grids are generated by independent heads, rollout steps, or post-processing, their differences need not correspond to evaluations of a common thermal evolution.
Moreover, simply resizing a prediction can change its display grid without informing the model of the requested sampling density, and pointwise supervision at hourly base-grid targets does not constrain non-integer lead queries or cross-resolution agreement.
Our key insight is that query flexibility must be coupled with state sharing and consistency regularization.
Accordingly, \method{} uses three coordinated design choices: a shared latent meteorological state, an explicit query-conditioned coordinate decoder, and coherence-preserving objectives for space, time, and scale.
As illustrated in Fig.~\ref{fig:framework}, the framework follows the mapping
$\bX \xrightarrow{\mathcal{E}_{\theta}} \bZ \xrightarrow{\mathcal{D}_{\theta}(\tau,\bP_{H_o,W_o},\bR_{H_o,W_o})} \hY_{\tau}^{(H_o,W_o)}$.
Specifically, a multi-scale meteorological encoder transforms the multi-variable \era{} history $\bX$ into a latent regional weather state $\bZ$.
The requested lead time $\tau$, output coordinates $\bP_{H_o,W_o}$, and grid resolution $\bR_{H_o,W_o}$ then condition a coordinate decoder, so each forecast product is obtained by evaluating the same state under a specified query.
Training combines deterministic, spatial-gradient, temporal-difference, and scale-consistency objectives to anchor supervised accuracy while regularizing regional thermal contrast, lead-wise evolution, and cross-resolution agreement.

\subsection{Query-conditioned field model}

We formalize \method{} as a neural field conditioned on both the encoded atmospheric state and the requested forecast product.
This shared-state formulation prevents the query interface from degenerating into separate product-specific mappings.
The encoder first summarizes the input history into a shared latent regional state, after which the decoder evaluates this state at each requested lead time and grid cell:
\begin{equation}
  \begin{aligned}
  \bZ &= \mathcal{E}_{\theta}\!\left(\bX\right),\\
  \hY_{\tau}^{(H_o,W_o)}(i,j)
  &=
  \mathcal{D}_{\theta}\!\left(
    \bZ_{H_o,W_o}(i,j),
    \bphi(\tau),
    \bp_{i,j},
    \br_{H_o,W_o}
  \right),
  \end{aligned}
  \label{eq:cstf}
\end{equation}
where $\bZ$ denotes the latent regional weather state, $\bZ_{H_o,W_o}$ is its interpolation on the requested grid, $\bphi(\tau)$ is a sinusoidal lead-time embedding, $\bp_{i,j}\in[-1,1]^2$ is the normalized coordinate of grid cell $(i,j)$, and $\br_{H_o,W_o}=(1/H_o,1/W_o)$ encodes the requested sampling density.
This formulation makes the output grid an explicit part of the decoder input, enabling different spatial discretizations to be treated as direct evaluations of the same atmospheric state.

\subsection{Multi-scale meteorological encoder}

The encoder serves as the state-construction component of \method{}, rather than a product-specific prediction head.
Because this state is later queried across lead times and resolutions, it must preserve regional thermal gradients while aggregating broader wind, pressure, moisture, cloud, and radiation cues from the multi-variable \era{} history.
We instantiate this state encoder as an encoder--decoder backbone with skip aggregation, leaving $\bZ$ independent of any fixed lead time or output resolution and available for query-specific field evaluation by the coordinate decoder.

\subsection{Lead-time and resolution queries}

A forecast query in \method{} specifies when the temperature field is evaluated, where it is sampled, and at what output density it is requested.
For the temporal component, \method{} normalizes the requested lead $\tau$ by the maximum training horizon and maps it to a learnable temporal code:
\begin{equation}
  \bphi(\tau)
  =
  \MLP\!\left(
  \SinCos\!\left(\frac{\tau}{\tau_{\max}}\right)
  \right).
\end{equation}
The sinusoidal basis provides a smooth temporal representation, and the MLP adapts it to short-range temperature evolution.
The same embedding function is used for supervised integer leads and non-integer diagnostic queries such as $0.5$ h or $4.5$ h.

For the spatial component, coordinates alone identify where to evaluate the field, but they do not indicate whether the field is being sampled sparsely or densely.
We therefore encode the requested grid by both sampling locations and sampling density.
The first field is the normalized coordinate grid
\begin{equation}
  \bP_{H_o,W_o}
  =
  \bigl\{\bp_{i,j}=(x_j,y_i): x_j,y_i\in[-1,1]\bigr\},
\end{equation}
which specifies the location of each output cell within the regional domain.
The resolution field is constant over the output image:
\begin{equation}
  \bR_{H_o,W_o}(i,j)
  =
  \left(1/H_o,1/W_o\right).
\end{equation}
Together, $\bphi(\tau)$, $\bP_{H_o,W_o}$, and $\bR_{H_o,W_o}$ define the requested forecast product passed to the decoder.

\subsection{Coordinate decoder}

The coordinate decoder evaluates the shared weather state under the requested product query.
A conventional convolutional prediction head maps latent features to a fixed grid and is unaware of absolute sampling location or requested grid density, making resolution changes indistinguishable from post-hoc resizing.
We instead interpolate $\bZ$ to the requested grid and condition the decoder on local latent features, spatial coordinates, resolution code, and lead-time embedding:
\begin{equation}
  \hY_{\tau}^{(H_o,W_o)}
  =
  \mathcal{D}_{\theta}\!\left[
  \bZ_{H_o,W_o},
  \bP_{H_o,W_o},
  \bR_{H_o,W_o},
  \bphi(\tau)
  \right].
\end{equation}
In implementation, $\mathcal{D}_{\theta}$ is a lightweight convolutional decoder applied on the requested grid.
Because query channels are part of the decoder input, each predicted value depends on local meteorological features, spatial position, forecast lead time, and sampling scale.
Thus, outputs such as $96\times96$, $128\times128$, and $192\times192$ fields are generated by direct field evaluation rather than by resizing a fixed-grid forecast.

\begin{table*}[t]
  \centering
  \caption{
  Lead-wise quantitative comparison for short-range \tmtwo{} forecasting over Southeast China.
  RMSE and MAE are reported in degrees Celsius at each supervised forecast lead, where lower values indicate better accuracy.
  The best result in each column is highlighted in \textcolor{green!60!black}{green}, and the second-best result is underlined.
  }
  \label{tab:leadwise_skill}
  \setlength{\tabcolsep}{4.5pt}
  \renewcommand{\arraystretch}{1.08}
  \resizebox{\textwidth}{!}{
  \begin{tabular}{lcccccccccccc}
    \toprule
    \multirow{2}{*}{Model}
    & \multicolumn{2}{c}{t+6 h}
    & \multicolumn{2}{c}{t+5 h}
    & \multicolumn{2}{c}{t+4 h}
    & \multicolumn{2}{c}{t+3 h}
    & \multicolumn{2}{c}{t+2 h}
    & \multicolumn{2}{c}{t+1 h}\\
    \cmidrule(lr){2-3}
    \cmidrule(lr){4-5}
    \cmidrule(lr){6-7}
    \cmidrule(lr){8-9}
    \cmidrule(lr){10-11}
    \cmidrule(lr){12-13}
    & RMSE & MAE
    & RMSE & MAE
    & RMSE & MAE
    & RMSE & MAE
    & RMSE & MAE
    & RMSE & MAE \\
    \midrule
    \textit{Baseline}
    & 8.196 & 3.741
    & 7.992 & 3.378
    & 7.782 & 2.968
    & 7.583 & 2.529
    & 7.419 & 2.108
    & 7.309 & 1.779
\\

    SMAAT-UNet~\citep{trebing2021smaat}
    & 1.989 & 1.425
    & 1.785 & 1.270
    & 1.581 & 1.141
    & 1.347 & 1.000
    & 1.106 & 0.840
    & 1.000 & 0.765
 \\
 
    EarthFormer~\citep{gao2022earthformer}
    & 2.784 & 1.254
    & 2.668 & 1.154
    & 2.600 & 1.064
    & 2.590 & 0.990
    & 2.600 & 0.939
    & 2.615 & 0.902
\\

    ARROW~\citep{tian2025arrow}
    & 1.625 & 1.131
    & \second{1.444} & 1.015
    & \second{1.246} & \second{0.892}
    & \second{1.051} & \second{0.768}
    & \best{0.870} & \best{0.655}
    & \best{0.739} & \best{0.567}
\\

    Weather-RF~\citep{schusterbauer2026probabilistic}
    & \second{1.609} & \second{1.112}
    & 1.446 & \second{1.008}
    & 1.284 & 0.908
    & 1.126 & 0.812
    & 0.970 & 0.720
    & 0.853 & 0.647
\\

    Ours
    & \best{1.414} & \best{1.014}
    & \best{1.284} & \best{0.925}
    & \best{1.162} & \best{0.845}
    & \best{1.036} & \best{0.765}
    & \second{0.909} & \second{0.685}
    & \second{0.823} & \second{0.631}
\\
    \bottomrule
  \end{tabular}
  }
\end{table*}

\subsection{Training objective}

Training uses hourly lead times $\{\tau_k\}_{k=1}^{T_{\mathrm{out}}}$ with available \eraland{} references.
The supervised forecast loss anchors \method{} to observed \eraland{} targets:
\begin{equation}
  \mathcal{L}_{\mathrm{forecast}}
  =
  \frac{1}{T_{\mathrm{out}}}
  \sum_{k=1}^{T_{\mathrm{out}}}
  \operatorname{MSE}\!\left(
  \hY_{\tau_k}^{(H,W)},
  \bY_{\tau_k}^{(H,W)}
  \right).
\end{equation}
However, pointwise hourly supervision alone does not guarantee coherent behavior when the same field evaluator is queried across resolutions or adjacent lead times.
We therefore add three regularizers for scale agreement, regional thermal structure, and lead-wise evolution.
The scale-consistency loss matches a direct low-resolution query with an area-downsampled base-grid query, encouraging different resolution evaluations of the same atmospheric state to agree after aggregation:
\begin{equation}
  \begin{aligned}
  \mathcal{L}_{\mathrm{scale}}
  =
  \frac{1}{T_{\mathrm{out}}}
  \sum_{k=1}^{T_{\mathrm{out}}}
  \operatorname{MSE}\!\Bigl(
  &\mathcal{F}_{\theta}\!\left(\bX;\tau_k,H/2,W/2\right),\\
  &\operatorname{AreaDown}
  \left(\mathcal{F}_{\theta}\!\left(\bX;\tau_k,H,W\right)\right)
  \Bigr),
  \end{aligned}
\end{equation}
where $\operatorname{AreaDown}$ denotes area downsampling from $(H,W)$ to $(H/2,W/2)$.

The spatial-gradient loss discourages oversmoothing by matching finite-difference \tmtwo{} gradients:
\begin{equation}
  \begin{aligned}
  \mathcal{L}_{\mathrm{grad}}
  =
  \frac{1}{T_{\mathrm{out}}}
  \sum_{k=1}^{T_{\mathrm{out}}}
  \Bigl(
  &\left\|\nabla_x \hY_{\tau_k}
  -\nabla_x \bY_{\tau_k}\right\|_1\\
  +
  &\left\|\nabla_y \hY_{\tau_k}
  -\nabla_y \bY_{\tau_k}\right\|_1
  \Bigr).
  \end{aligned}
\end{equation}

\begin{table}[htbp]
  \centering
  \caption{
  Overall quantitative comparison for integer-lead \tmtwo{} forecasting over Southeast China.
  Lower is better for RMSE, MAE, and bias magnitude, while higher is better for correlation.
  }
  \label{tab:main_skill}
  \resizebox{0.5\textwidth}{!}{
  \begin{tabular}{lcccc}
    \toprule
    Model & RMSE $\downarrow$ & MAE $\downarrow$ & Bias $\downarrow$ & Corr. $\uparrow$ \\
    \midrule
    \textit{Baseline} & 7.720 & 2.751 & 1.247 & 0.576 \\
    SMAAT-UNet~\citep{trebing2021smaat} & 1.510 & 1.074 & \second{0.329} & 0.987 \\
    EarthFormer~\citep{gao2022earthformer} & 2.644 & 1.051 & 0.598 & 0.963 \\
    ARROW~\citep{tian2025arrow} & \second{1.203} & \second{0.838} & 0.389 & \second{0.992} \\
    Weather-RF~\citep{schusterbauer2026probabilistic} & 1.243 & 0.868 & 0.446 & \second{0.992} \\
    Ours & \best{1.123} & \best{0.811} & \best{0.273} & \best{0.993} \\
    \midrule
    \textit{Improv. vs. Second}
    & \cellcolor{gray!20}\textcolor{red}{+6.7\%}
    & \cellcolor{gray!20}\textcolor{red}{+3.2\%}
    & \cellcolor{gray!20}\textcolor{red}{+17.0\%}
    & \cellcolor{gray!20}\textcolor{red}{+0.1\%} \\
    \bottomrule
  \end{tabular}}
\end{table}

The temporal-difference loss constrains short-range evolution by aligning frame-to-frame temperature changes:
\begin{equation}
  \begin{aligned}
  \mathcal{L}_{\mathrm{temp}}
  =
  \frac{1}{T_{\mathrm{out}}-1}
  \sum_{k=1}^{T_{\mathrm{out}}-1}
  \Bigl\|
  \left(\hY_{\tau_{k+1}}-\hY_{\tau_k}\right)\\
  -
  \left(\bY_{\tau_{k+1}}-\bY_{\tau_k}\right)
  \Bigr\|_1.
  \end{aligned}
\end{equation}
The overall training objective is
\begin{equation}
  \mathcal{L}
  =
  \mathcal{L}_{\mathrm{forecast}}
  +
  \lambda_g \mathcal{L}_{\mathrm{grad}}
  +
  \lambda_t \mathcal{L}_{\mathrm{temp}}
  +
  \lambda_s \mathcal{L}_{\mathrm{scale}}.
\end{equation}

\subsection{Inference and diagnostic queries}

After training, inference is carried out by querying the same evaluator $\mathcal{F}_{\theta}$ with specified lead-time and grid-size arguments.
Integer leads are verified against \eraland{} using deterministic forecast metrics.
Non-integer leads are reported as forecast-only diagnostics because the current hourly target split has no direct half-hour references.
For resolution-controllable inference, predictions at multiple output sizes are downsampled to a common base grid to quantify scale consistency.
This diagnostic does not replace verification against \eraland{}; it tests whether the query-conditioned decoder returns mutually coherent fields when the same atmospheric state is sampled at different spatial resolutions.

\section{Experiments}
\label{sec:experiments}

\subsection{Implementation Details}
\noindent\textbf{Experimental Setup.}
We evaluate \method{} on short-range 2-m temperature forecasting from \era{} to \eraland{}.
Each sample uses a three-hour, nine-variable \era{} history to predict six hourly future \eraland{} \tmtwo{} fields.
The main benchmark is conducted over Southeast China at a base grid of $128\times128$, and a global-scope split is used only for qualitative query diagnostics.
Detailed data splits, preprocessing, optimization, objective weights, checkpoint selection, and query-evaluation protocols are provided in Appendix.

\noindent\textbf{Evaluation Metrics.}
Following meteorological-state evaluation protocols \citep{tu2025satellite}, we report MAE and RMSE in physical units for supervised integer leads.
We also use bias and spatial correlation for regional temperature diagnostics, and cross-resolution scale-consistency error for resolution-controllable prediction.
We define the reported scale-consistency error in normalized model space as
\begin{equation}
  \mathcal{E}_{\mathrm{scale}}(r)
  =
  \operatorname{RMSE}\!\left(
  \operatorname{Resample}^{\mathrm{area}}_{r\rightarrow r_0}
  \left(
  \hY_{\tau}^{(r,r)}
  \right),
  \hY_{\tau}^{(r_0,r_0)}
  \right),
\end{equation}
where $r_0=128$ is the base resolution and $\operatorname{Resample}^{\mathrm{area}}_{r\rightarrow r_0}$ maps each query to the base grid.
This diagnostic compares predictions across query grids rather than measuring physical-unit error against \eraland{} references.

\noindent\textbf{Baseline Comparisons.}
We compare \method{} with baseline, SMAAT-UNet \citep{trebing2021smaat}, EarthFormer \citep{gao2022earthformer}, ARROW-style forecasting \citep{tian2025arrow}, and Weather-RF/FREUD-style forecasting \citep{schusterbauer2026probabilistic}.
All learned baselines follow the same data, input--output, normalization, and evaluation protocol; controlled \method{} variants only remove selected objective terms.

\begin{figure}[ht]
  \centering
  \includegraphics[width=\linewidth]{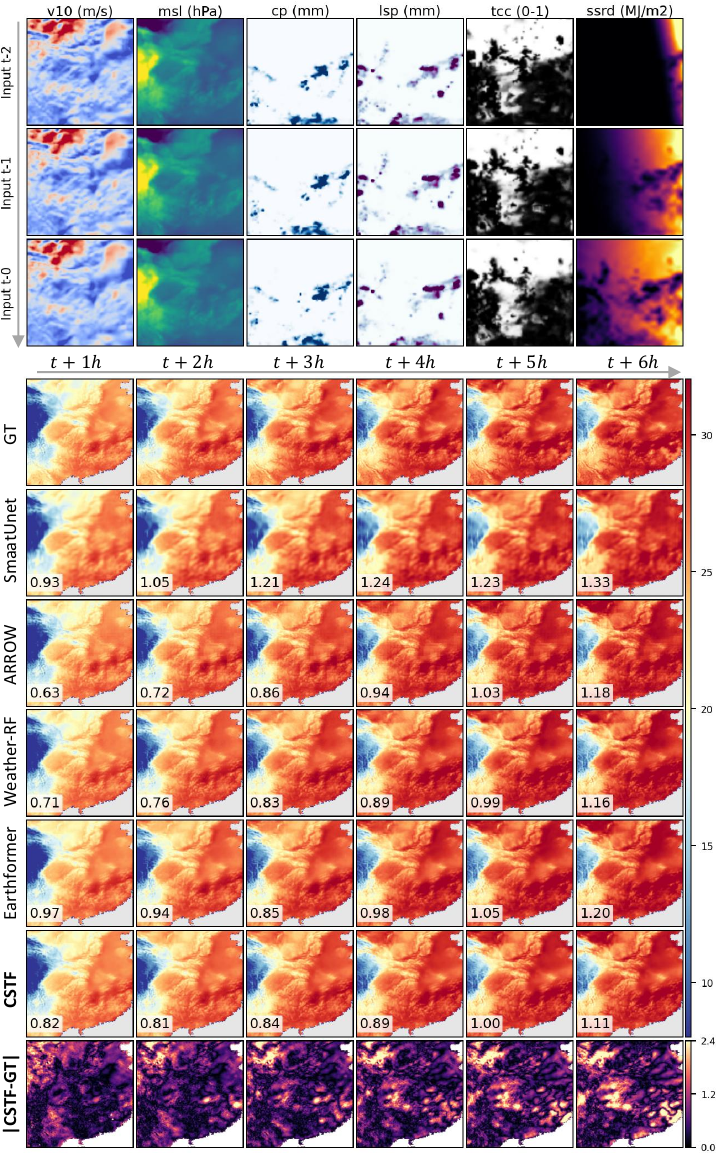}
  \caption{
  Qualitative comparison of deterministic \tmtwo{} forecasts over the Southeast China benchmark.
  Panels show the \era{} input context, integer-lead \method{} predictions, \eraland{} references, and absolute-error maps.
  }
  \label{fig:regional_case}
\end{figure}
\subsection{Regional deterministic forecast}

Before analyzing flexible query behavior, we first verify the deterministic forecasting skill of \method{} at the supervised hourly leads on the Southeast China benchmark.

Table~\ref{tab:leadwise_skill} reports the lead-wise comparison.
Although ARROW attains slightly lower errors at the first two leads, \method{} achieves the best skill from t+3 h to t+6 h, indicating more stable performance as short-range evolution becomes less constrained by the input state.

The aggregate results in Table~\ref{tab:main_skill} further show that \method{} obtains the lowest average RMSE, MAE, and bias, together with the highest spatial correlation among the compared methods.
Relative to the strongest fixed-grid baseline, ARROW-style forecasting, \method{} reduces RMSE from $1.203^\circ$C to $1.123^\circ$C and MAE from $0.838^\circ$C to $0.811^\circ$C.
The representative case in Fig.~\ref{fig:regional_case} shows that this improvement is accompanied by coherent regional thermal structures, with larger errors mainly confined to localized areas.

\begin{figure}[ht]
  \centering
  \includegraphics[width=\linewidth]{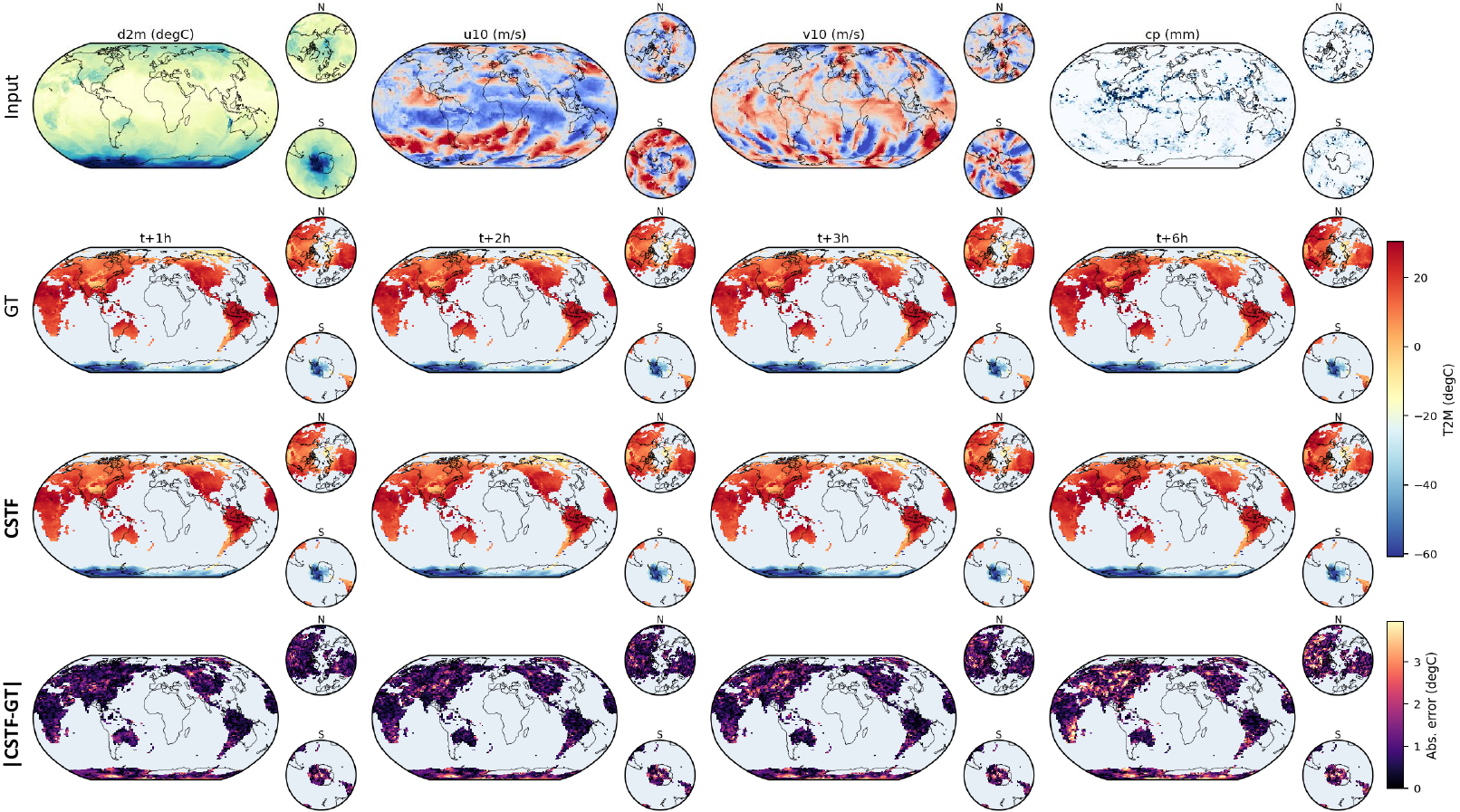}
  \caption{
  Global-scope diagnostic visualization of \tmtwo{} field prediction.
  Panels compare \eraland{} references, \method{} predictions, and absolute-error maps under a Robinson projection with polar insets.
  Zoomed-in views are provided in Appendix.
  }
  \label{fig:global_diag}
\end{figure}
\begin{table*}[t]
  \centering
  \caption{
  Extended lead-wise ablation of \method{} objective terms on the Southeast China test split.
  RMSE and MAE are reported in degrees Celsius at each supervised hourly lead; lower is better.
  This table extends the main-text lead-wise ablation by separately removing the spatial-gradient and temporal-difference regularizers.
    The best score in each column is highlighted in \textcolor{green!60!black}{green}, and the second-best score is underlined.
  }
  \label{tab:app_extended_ablation_leadwise}
  \setlength{\tabcolsep}{4.2pt}
  \renewcommand{\arraystretch}{1.08}
  \resizebox{\textwidth}{!}{%
  \begin{tabular}{lcccccccccccccc}
    \toprule
    \multirow{2}{*}{Variant}
    & \multicolumn{2}{c}{t+6 h}
    & \multicolumn{2}{c}{t+5 h}
    & \multicolumn{2}{c}{t+4 h}
    & \multicolumn{2}{c}{t+3 h}
    & \multicolumn{2}{c}{t+2 h}
    & \multicolumn{2}{c}{t+1 h}
    & \multicolumn{2}{c}{Mean} \\
    \cmidrule(lr){2-3}
    \cmidrule(lr){4-5}
    \cmidrule(lr){6-7}
    \cmidrule(lr){8-9}
    \cmidrule(lr){10-11}
    \cmidrule(lr){12-13}
    \cmidrule(lr){14-15}
    & RMSE & MAE
    & RMSE & MAE
    & RMSE & MAE
    & RMSE & MAE
    & RMSE & MAE
    & RMSE & MAE
    & RMSE & MAE \\
    \midrule

    w/o aux. losses
    & 1.555 & 1.085
    & 1.403 & 0.984
    & 1.248 & 0.890
    & 1.096 & 0.800
    & 0.960 & 0.720
    & 0.871 & 0.668
    & 1.213 & 0.858 \\

    w/o $\mathcal{L}_{\mathrm{scale}}$
    & 1.575 & 1.096
    & 1.422 & 0.996
    & 1.263 & 0.898
    & 1.100 & 0.802
    & 0.949 & 0.712
    & 0.852 & 0.654
    & 1.220 & 0.860 \\

    w/o $\mathcal{L}_{\mathrm{grad}}$
    & 1.581 & 1.106
    & 1.420 & 1.003
    & 1.264 & 0.907
    & 1.116 & 0.817
    & 0.974 & 0.730
    & 0.875 & 0.669
    & 1.230 & 0.872 \\

    w/o $\mathcal{L}_{\mathrm{temp}}$
    & \second{1.476} & \second{1.043}
    & \second{1.337} & \second{0.948}
    & \second{1.203} & \second{0.861}
    & \second{1.059} & \second{0.773}
    & \second{0.921} & \second{0.689}
    & \second{0.831} & \second{0.636}
    & \second{1.160} & \second{0.825} \\

    Ours
    & \best{1.414} & \best{1.014}
    & \best{1.284} & \best{0.925}
    & \best{1.162} & \best{0.845}
    & \best{1.036} & \best{0.765}
    & \best{0.909} & \best{0.685}
    & \best{0.823} & \best{0.631}
    & \best{1.123} & \best{0.811} \\

    \bottomrule
  \end{tabular}%
  }
\end{table*}

\subsection{Global diagnostic visualization}

To assess whether the query formulation preserves coherent large-scale thermal structures beyond the regional benchmark, we further apply \method{} to the global-scope split as a qualitative diagnostic.
Figure~\ref{fig:global_diag} presents the input context, \method{} forecasts, \eraland{} references, and absolute-error maps under a Robinson projection with polar insets.
This visualization is used to inspect large-scale thermal coherence and spatial error patterns, while all quantitative comparisons are still computed on the native model grid.

\subsection{Continuous lead-time query}

\begin{figure}[ht]
  \centering
  \includegraphics[width=\linewidth]{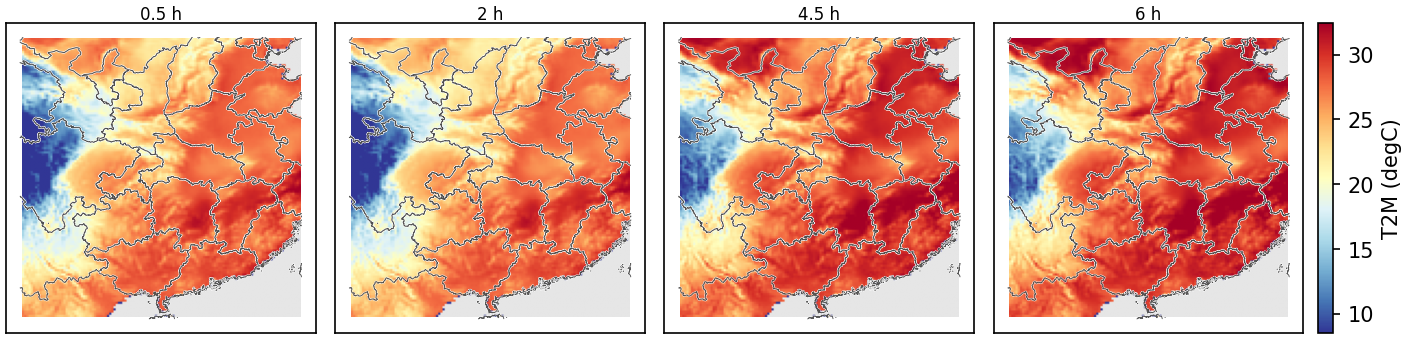}
  \caption{
  Continuous lead-time query.
  The same trained model is queried at $0.5$, $2$, $4.5$, and $6$ hours.
  The $0.5$ h and $4.5$ h panels are non-integer forecast-only queries, while the $2$ h and $6$ h panels align with supervised hourly targets.
  }
  \label{fig:lead_query}
\end{figure}

\begin{figure}[ht]
  \centering
  \includegraphics[width=\linewidth]{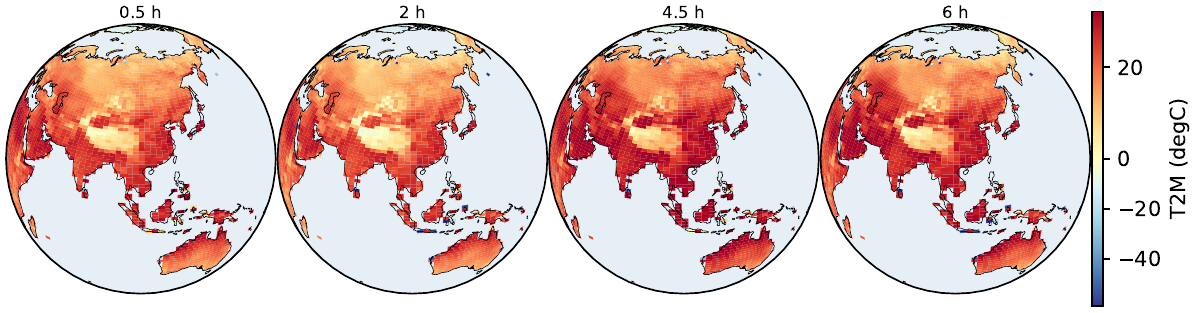}
  \caption{
  Global continuous lead-time query on a Robinson projection with polar insets.
  The same checkpoint is queried at $0.5$, $2$, $4.5$, and $6$ hours, using the same lead-time set as the regional query visualization.
  }
  \label{fig:global_lead_query}
\end{figure}

Because \method{} represents lead time as an explicit query variable, we test whether the learned field evaluator can produce temporally ordered forecasts beyond the hourly supervision grid.
The temporal queries include both verified integer leads and diagnostic non-integer leads:
\begin{equation}
  \tau\in\{0.5,2,4.5,6\}\ \mathrm{hours}.
\end{equation}
The integer leads are compared with \eraland{} targets, whereas the half-hour queries are reported as forecast-only diagnostics because the target split is hourly.
Figure~\ref{fig:lead_query} shows that regional temperature structures evolve smoothly across the queried lead times, and Fig.~\ref{fig:global_lead_query} provides the corresponding global diagnostic view.

\subsection{Resolution-controllable prediction}

\begin{figure}[ht]
  \centering
  \includegraphics[width=\linewidth]{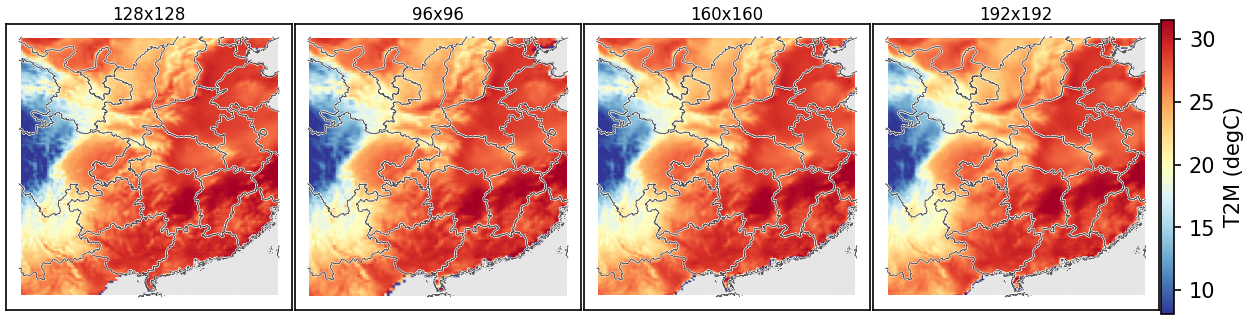}
  \caption{
  Resolution-controllable prediction.
  Fixed input and lead-time query are decoded at $96\times96$, $128\times128$, $160\times160$, and $192\times192$ grids.
  Panels show direct resolution-conditioned decoding rather than post-hoc resizing.
  }
  \vspace{-10pt}
  \label{fig:resolution_query}
\end{figure}

\begin{figure}[ht]
  \centering
  \includegraphics[width=\linewidth]{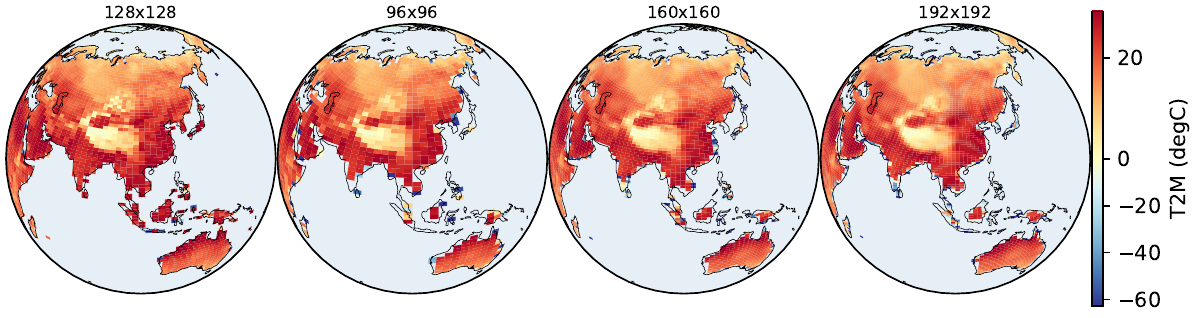}
  \caption{
  Global resolution-controllable prediction.
  Orthographic visualizations are generated for different output grid sizes from the same input state and lead-time query.
  }
  \vspace{-10pt}
  \label{fig:global_resolution_query}
\end{figure}

We then assess spatial queryability by evaluating the learned field decoder on output grids that differ from the base training resolution:
\begin{equation}
  (H_o,W_o)\in
  \{96\times96,128\times128,160\times160,192\times192\}.
\end{equation}
Figures~\ref{fig:resolution_query} and \ref{fig:global_resolution_query} show that \method{} directly decodes temperature fields at the requested grids, rather than resizing a fixed-grid prediction.
The resulting panels preserve the main thermal patterns across query resolutions, supporting the intended resolution-controllable inference behavior.

\subsection{Scale-consistency analysis}

Resolution-controllable visualization alone does not ensure compatible temperature states across grids.
We therefore aggregate non-base queries to the reference grid and compare them with the base-grid prediction.
The resulting diagnostic, reported in Table~\ref{tab:scale_consistency}, isolates the resolution-query training signal under controlled ablations.

\begin{table}[t]
  \centering
  \caption{
  Scale-consistency ablation for resolution-controllable prediction.
  Values are normalized cross-resolution RMSE after resampling to $128\times128$; lower is better.
  Best results are highlighted in \textcolor{green!60!black}{green}, and second-best results are underlined.
  }
  \label{tab:scale_consistency}
  \resizebox{0.5\textwidth}{!}{%
  \begin{tabular}{lcccc}
    \toprule
    Variant & $96\rightarrow128$ & $160\rightarrow128$ & $192\rightarrow128$ & Mean \\
    \midrule
    w/o aux. losses & 1.396 & 1.165 & 1.233 & 1.265 \\
    w/o $\mathcal{L}_{\mathrm{scale}}$ & 1.449 & 1.171 & 1.266 & 1.295 \\
    w/o $\mathcal{L}_{\mathrm{grad}}$ & 1.111 & \second{1.082} & 0.997 & \second{1.060} \\
    w/o $\mathcal{L}_{\mathrm{temp}}$ & \second{1.099} & 1.131 & \second{0.988} & 1.072\\
    Ours & \best{1.098} & \best{1.077} & \best{0.982} & \best{1.052} \\
    \bottomrule
  \end{tabular}}
  \vspace{-10pt}
\end{table}

\begin{table}[t]
  \centering
  \caption{
  Objective-term ablation on the Southeast China test split.
  All variants use the same nine \era{} input variables and architecture.
  Scale error is the mean normalized cross-resolution RMSE after resampling non-base queries to $128\times128$.
  Best results are highlighted in \textcolor{green!60!black}{green}, and second-best results are underlined.
  }
  \label{tab:ablation_results}
  \resizebox{0.5\textwidth}{!}{%
  \begin{tabular}{lccccc}
    \toprule
    Variant & RMSE $\downarrow$ & MAE $\downarrow$ & Bias $\downarrow$ & Grad. err. $\downarrow$ & Scale err. $\downarrow$ \\
    \midrule
    w/o aux. losses & 1.213 & 0.858 & 0.339 & 0.468 & 1.265 \\
    w/o $\mathcal{L}_{\mathrm{scale}}$ & 1.220 & 0.860 & 0.350 & 0.382 & 1.295 \\
    w/o $\mathcal{L}_{\mathrm{grad}}$ & 1.230 & 0.872 & 0.356 & 0.455 & 1.060 \\
    w/o $\mathcal{L}_{\mathrm{temp}}$ & \second{1.160} & \second{0.825} & \second{0.281} & \second{0.376} & \best{1.038} \\
    Ours & \best{1.123} & \best{0.811} & \best{0.273} & \best{0.361} & \second{1.052} \\
    \bottomrule
  \end{tabular}%
  }
  \vspace{-10pt}
\end{table}

\subsection{Ablation Studies}

We evaluate two variants: w/o aux. losses keeps only $\mathcal{L}_{\mathrm{forecast}}$, while w/o $\mathcal{L}_{\mathrm{scale}}$ retains $\mathcal{L}_{\mathrm{grad}}$ and $\mathcal{L}_{\mathrm{temp}}$ but removes the scale-consistency term.
As shown in Table~\ref{tab:app_extended_ablation_leadwise}, the full objective improves RMSE and MAE across all supervised leads, with larger gains at longer horizons, indicating more stable lead-wise temperature evolution.
Table~\ref{tab:scale_consistency} shows that removing $\mathcal{L}_{\mathrm{scale}}$ increases cross-resolution error after aggregation, supporting its role in coherent resolution-query prediction.
Table~\ref{tab:ablation_results} further indicates that the full objective provides balanced gains in aggregate accuracy, thermal-structure fidelity, and scale coherence.

\clearpage
\section{Conclusion}

In this paper, we address the fixed-output limitation of regional \tmtwo{} forecasting by introducing \method{}, a spatiotemporal field forecaster.
Based on the view that temperature products at varying lead times and grids are evaluations of a shared evolving atmospheric state, \method{} encodes multi-variable \era{} histories and decodes \tmtwo{} fields with explicit lead-time, spatial-coordinate, and resolution queries.
The proposed deterministic, spatial-gradient, temporal-difference, and scale-consistency objectives promote forecast accuracy, regional thermal-structure fidelity, and cross-resolution coherence.
Experiments on the SE China benchmark show that \method{} improves fixed-lead deterministic skill while enabling non-integer lead-time and resolution-controllable queries.


\bibliographystyle{IEEEtran}
\bibliography{egbib}

\clearpage
\appendix
\label{appendix}

This supplementary material provides additional implementation details in Sec.~\ref{app:implementation}, extended temperature forecast case studies in Sec.~\ref{app:forecast_cases}, and query, training, benchmark, and ablation diagnostics in Sec.~\ref{app:visual_diagnostics}.

\subsection{Additional Implementation Details}
\label{app:implementation}

\noindent\textbf{Data split and preprocessing.}
The Southeast China (SE China) domain spans $100.0^\circ$E--$120.0^\circ$E and $20.0^\circ$N--$40.0^\circ$N.
The Southeast China and global-scope splits both contain 4,615 training samples, 511 validation samples, and 714 test samples.
The data are chronologically split over 2024--2025, with validation drawn from the training period using a 0.1 fraction and a held-out test subset used only for final evaluation.
Unless otherwise stated, fields are resized to $128\times128$ for training and supervised integer-lead evaluation.

\noindent\textbf{Training protocol.}
All reported \method{} models are trained for \tmtwo{} prediction with a three-frame input history and a six-frame hourly forecast target.
The main \method{} model is trained for 50 epochs with batch size 8, AdamW optimization, an initial learning rate of $10^{-3}$, weight decay of $10^{-4}$, cosine learning-rate annealing, gradient clipping at 1.0, and random seed 42.
The final model is selected by validation loss and then evaluated on the held-out test split.
For the full objective, the auxiliary loss weights are $\lambda_s=0.05$, $\lambda_g=0.05$, and $\lambda_t=0.01$.
Test metrics are computed with batch size 32 using the same normalization and evaluation script across \method{} and all fixed-grid baselines.

\begin{figure*}[ht]
  \centering
  \includegraphics[width=\textwidth]{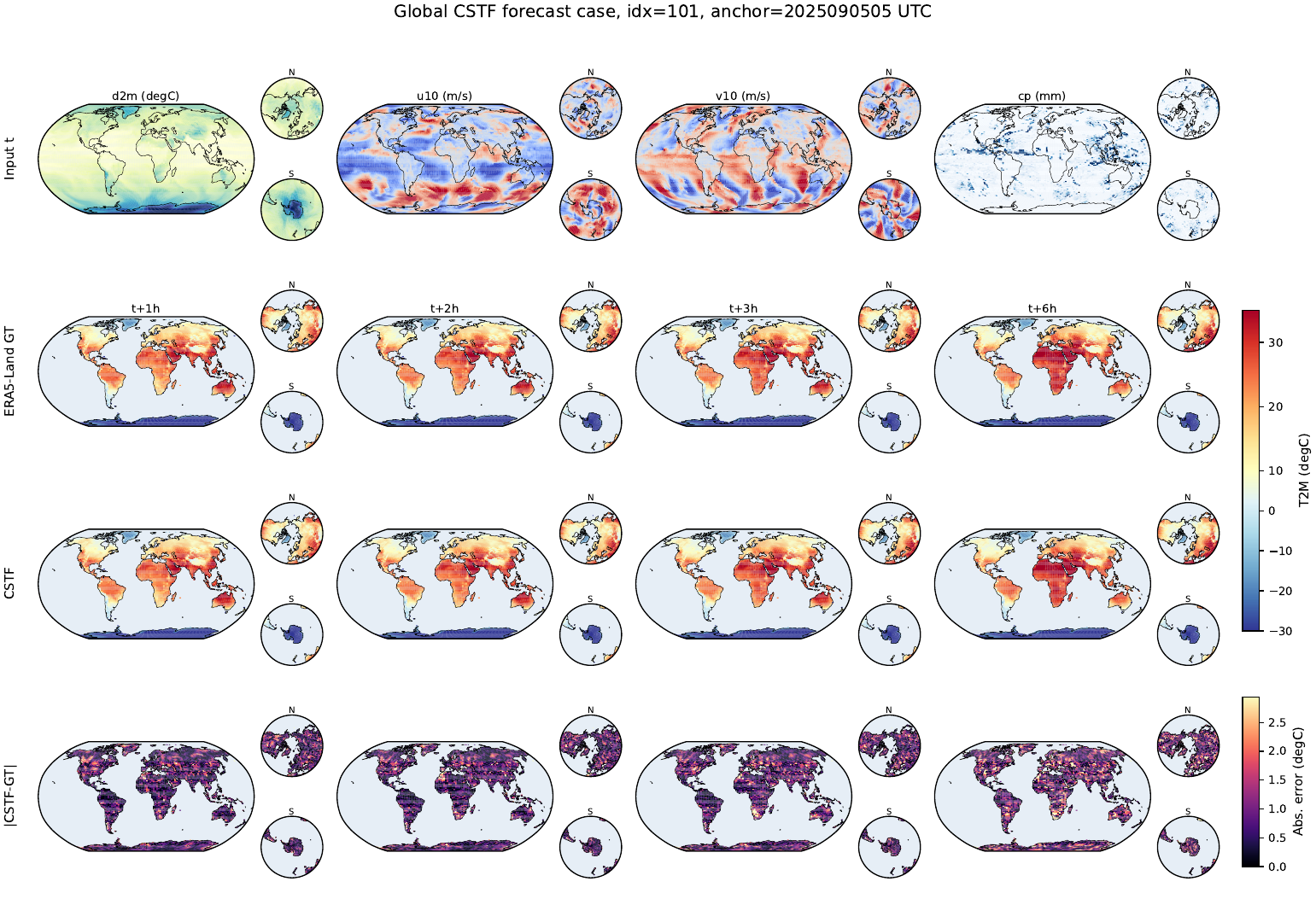}
  \caption{
  Additional deterministic forecast diagnostic for the global domain ($180.0^\circ$W--$180.0^\circ$E, $90.0^\circ$S--$90.0^\circ$N).
  The case shows input context, \eraland{} references, \method{} predictions, and absolute-error maps under the main-text global visualization protocol.
  It supplements the idx0 main-text case and assesses whether the qualitative global diagnostic remains consistent across forecast initializations.
  }
  \label{fig:app_global_diag_cases}
\end{figure*}

\noindent\textbf{Query diagnostics.}
The standard benchmark verifies supervised integer-lead \tmtwo{} forecasts against \eraland{}.
Continuous lead-time diagnostics use a single learned parameterization at both integer and non-integer lead times; non-integer leads are shown as forecast-only examples because the current hourly target split has no direct half-hour references.
Resolution diagnostics evaluate the same field parameterization at $96\times96$, $128\times128$, $160\times160$, and $192\times192$ output grids and assess cross-resolution consistency after resampling to the base grid.
The scale-consistency values in the main tables are reported as normalized RMSE diagnostics rather than physical-unit forecast errors.

\noindent\textbf{Ablation setting.}
All controlled \method{} variants keep the architecture, data split, input variables, input length, output horizon, training resolution, and model-selection rule fixed.
The ablations therefore isolate objective terms rather than changes in model capacity or input information.

\noindent\textbf{Training objective rationale.}
The forecast MSE provides the main deterministic supervision on the hourly \eraland{} targets and is computed over all supervised leads, channels, and grid cells in each training batch.
The scale-consistency term is designed for the queryable-resolution interface: for the same atmospheric input and lead time, a direct low-resolution query should agree with an area-downsampled high-resolution query after both are represented on the same grid.
In the implementation, the downsampled high-resolution branch is detached in this term, so the auxiliary gradient primarily aligns the direct low-resolution query with the high-resolution prediction used as the reference.
The spatial-gradient term compares finite differences along the two image axes and discourages overly smooth temperature maps by penalizing mismatches in local thermal contrasts.
The temporal-difference term compares adjacent-lead changes rather than only absolute values, encouraging the predicted sequence to follow the observed short-range warming or cooling tendency across the six supervised lead times.
Together, these auxiliary objectives are intended to support the two query diagnostics used in the experiments: coherent multi-resolution products and smooth intermediate lead-time behavior.

\begin{figure*}[ht]
  \centering
  \includegraphics[width=0.88\linewidth]{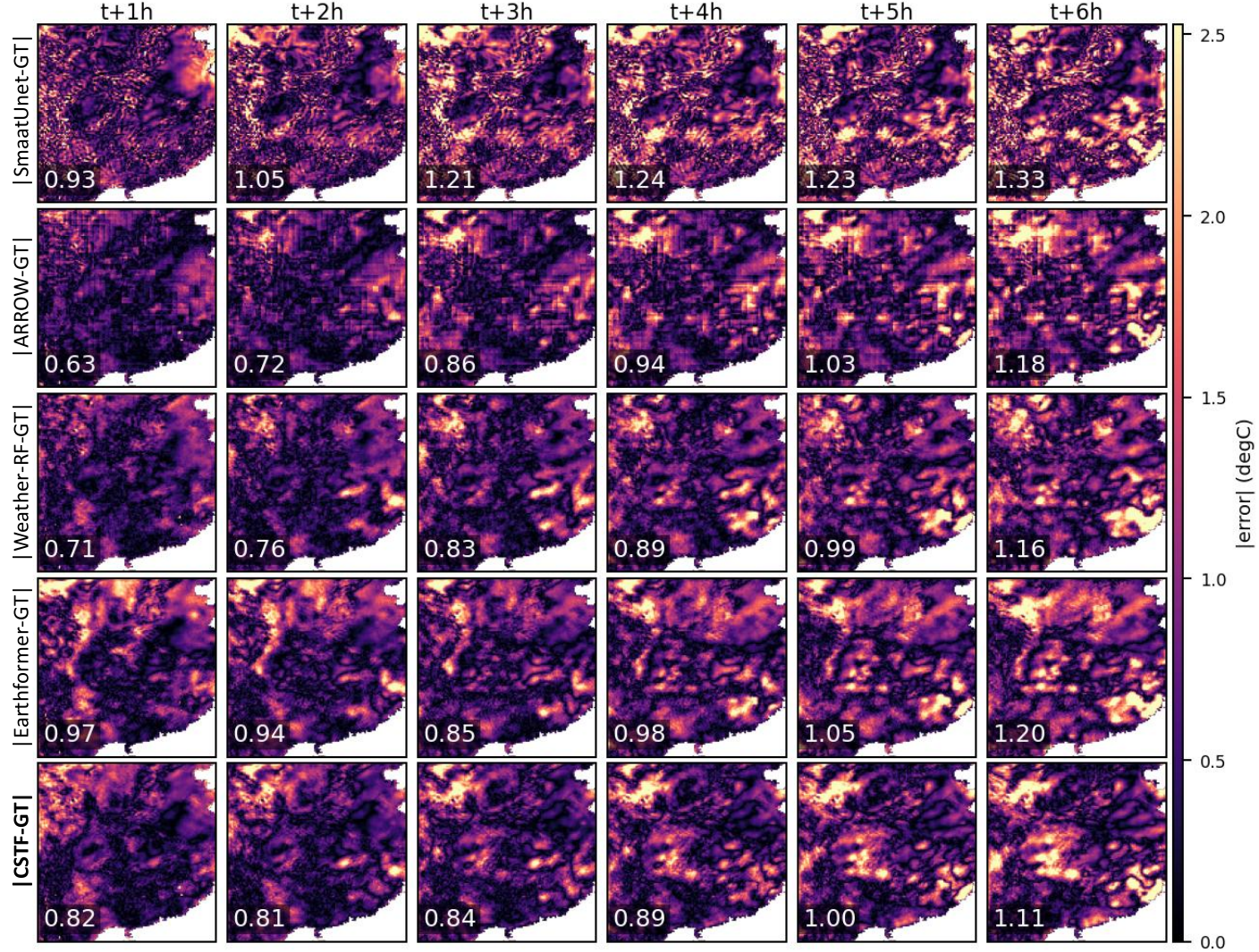}
  \caption{
  Lead-wise absolute-error maps for the Southeast China case study ($100.0^\circ$E--$120.0^\circ$E, $20.0^\circ$N--$40.0^\circ$N).
  Errors are shown over the six supervised integer lead times, providing a spatial diagnostic of where forecast discrepancies grow with lead time.
  The overlaid number on each panel reports the per-frame spatial RMSE ($^\circ$C) against the \eraland{} \tmtwo{} reference over valid land pixels.
  }
  \label{fig:regional_error}
\end{figure*}

\subsection{Additional Temperature Forecast Case Studies}
\label{app:forecast_cases}

This section provides additional deterministic \tmtwo{} forecast visualizations before the query-specific diagnostics in Sec.~\ref{app:visual_diagnostics}.
The purpose is to separate two forms of evidence: first, whether the learned forecaster produces meteorologically coherent temperature fields at the supervised hourly leads, and second, whether the same model can later be interrogated through lead-time and resolution queries.
Figures~\ref{fig:app_global_diag_cases} and \ref{fig:regional_error} therefore focus on standard deterministic forecast behavior rather than on flexible queryability.

Figure~\ref{fig:app_global_diag_cases} extends the global-scope qualitative diagnostic beyond the main-text initialization.
Because the global split is used only for diagnostic visualization, the figure should be interpreted as a spatial plausibility check: it examines whether large-scale thermal patterns, land--sea contrasts, and error distributions remain coherent when the regional field-evaluation formulation is applied under a global rendering protocol.
The inclusion of input context, \eraland{} references, \method{} predictions, and absolute-error maps allows the reader to distinguish forecast structure from residual error patterns.

Figure~\ref{fig:regional_error} complements the global case by focusing on the supervised Southeast China benchmark used for quantitative evaluation.
Instead of summarizing skill with a single aggregate score, the figure reports absolute-error maps at each integer lead time, making it possible to inspect where regional discrepancies emerge and how their spatial distribution changes across the 0--6 h forecast horizon.
The number overlaid on each panel is the corresponding spatial RMSE in degrees Celsius, computed between the model prediction and the \eraland{} \tmtwo{} reference over valid land pixels.
This view is particularly relevant for \tmtwo{} forecasting because localized thermal contrasts can be smoothed or displaced even when average error remains low.

\subsection{Additional Query and Quantitative Diagnostics}
\label{app:visual_diagnostics}

\begin{figure*}[t]
  \centering
  \includegraphics[width=0.98\textwidth]{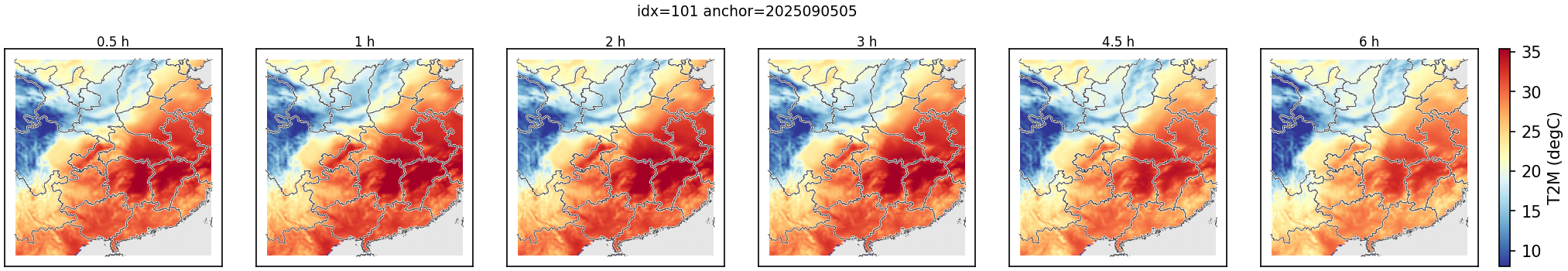}\\[2mm]
  \includegraphics[width=0.98\textwidth]{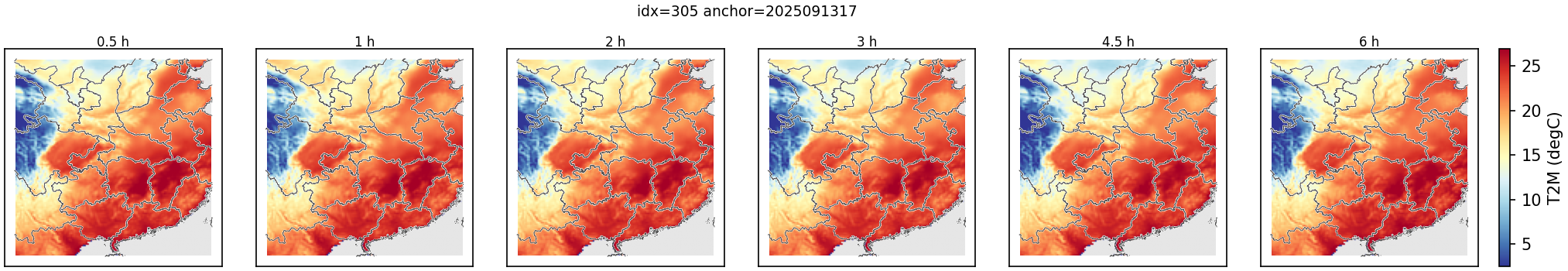}\\[2mm]
  \includegraphics[width=0.98\textwidth]{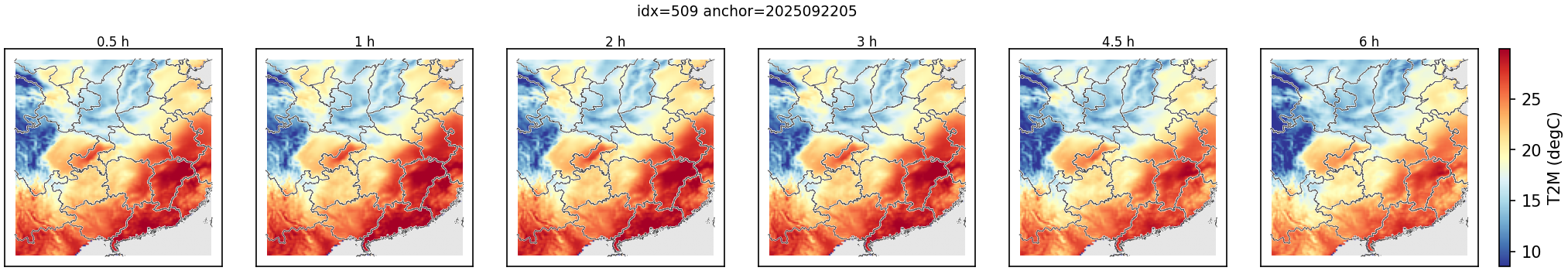}
  \caption{
  Continuous lead-time query examples over Southeast China ($100.0^\circ$E--$120.0^\circ$E, $20.0^\circ$N--$40.0^\circ$N).
  The three rows are independent test initializations at 2025-09-05 05:00 UTC, 2025-09-13 17:00 UTC, and 2025-09-22 05:00 UTC.
  Within each row, the same atmospheric input is queried at multiple lead times, including supervised hourly leads and non-integer lead times.
  The figure is therefore used to inspect whether the learned field evaluator produces temporally smooth and spatially coherent \tmtwo{} fields when the requested forecast time is varied continuously.
  Non-integer lead panels are forecast-only diagnostics because the current \eraland{} target sequence is available at hourly intervals.
  }
  \label{fig:app_lead_query_huadong_cases}
\end{figure*}

\begin{figure*}[t]
  \centering
  \includegraphics[width=0.98\textwidth]{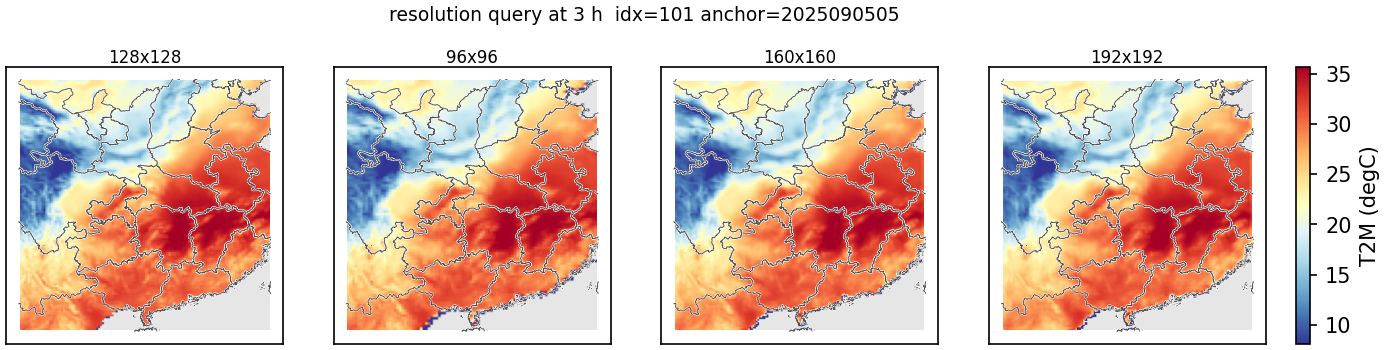}\\[2mm]
  \includegraphics[width=0.98\textwidth]{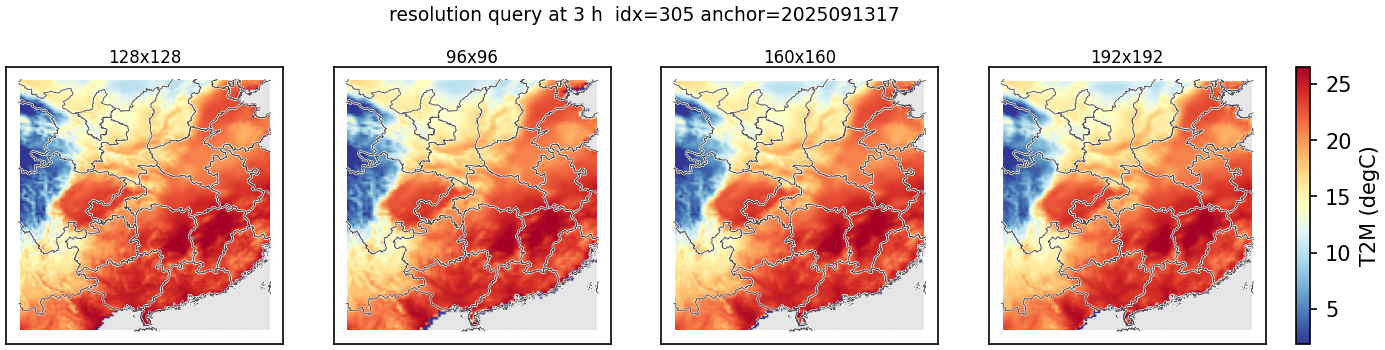}\\[2mm]
  \includegraphics[width=0.98\textwidth]{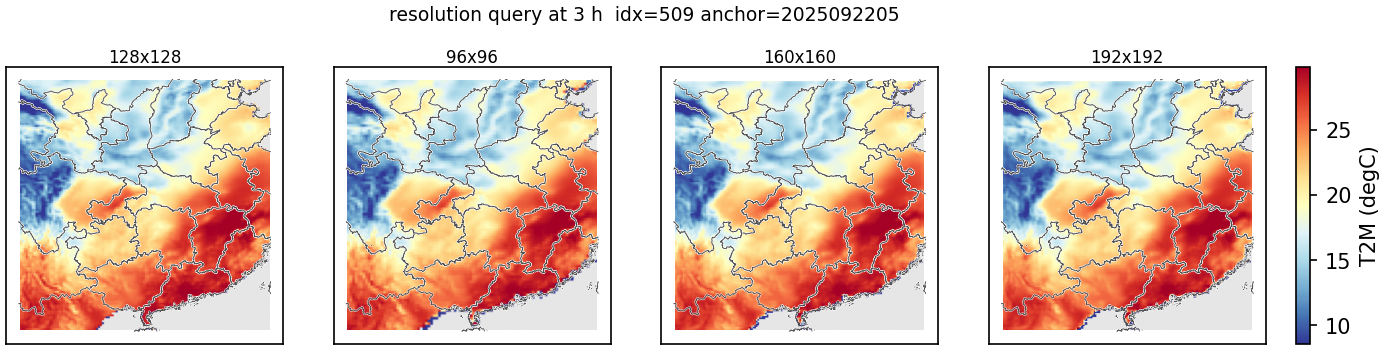}
  \caption{
  Resolution-conditioned query examples over Southeast China ($100.0^\circ$E--$120.0^\circ$E, $20.0^\circ$N--$40.0^\circ$N).
  The three rows are independent test initializations at 2025-09-05 05:00 UTC, 2025-09-13 17:00 UTC, and 2025-09-22 05:00 UTC.
  Within each row, \method{} directly evaluates the same forecast state at different requested output grids rather than resizing a single fixed-grid prediction.
  The visual comparison checks whether regional thermal structures remain consistent as the output resolution changes, complementing the cross-resolution RMSE values reported in the main text.
  }
  \label{fig:app_resolution_query_huadong_cases}
\end{figure*}

\begin{figure*}[t]
  \centering
  \includegraphics[width=0.98\textwidth]{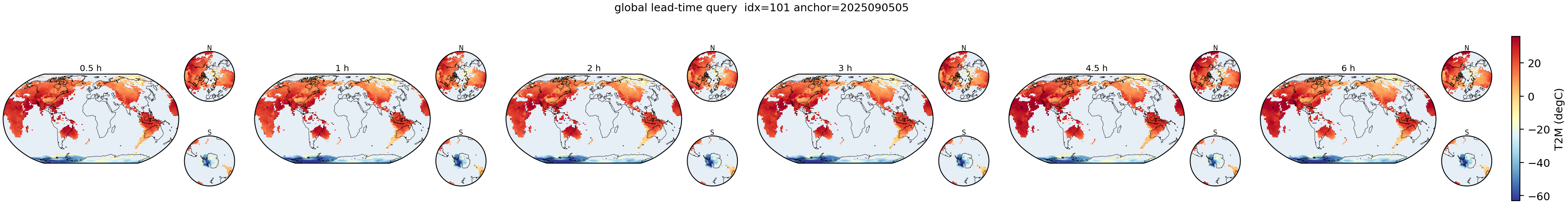}\\[2mm]
  \includegraphics[width=0.98\textwidth]{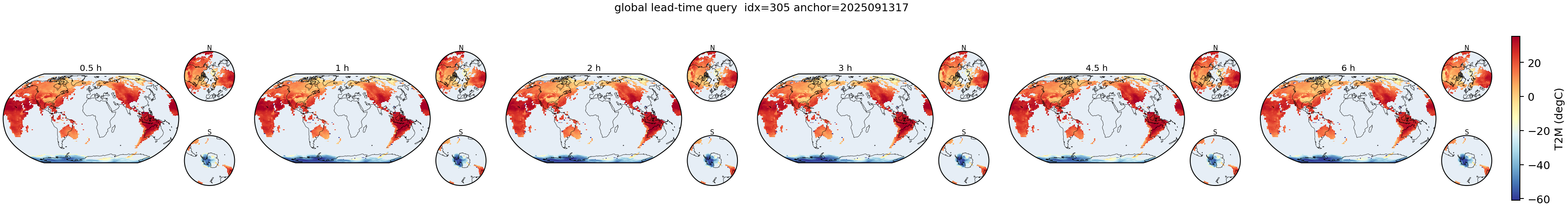}\\[2mm]
  \includegraphics[width=0.98\textwidth]{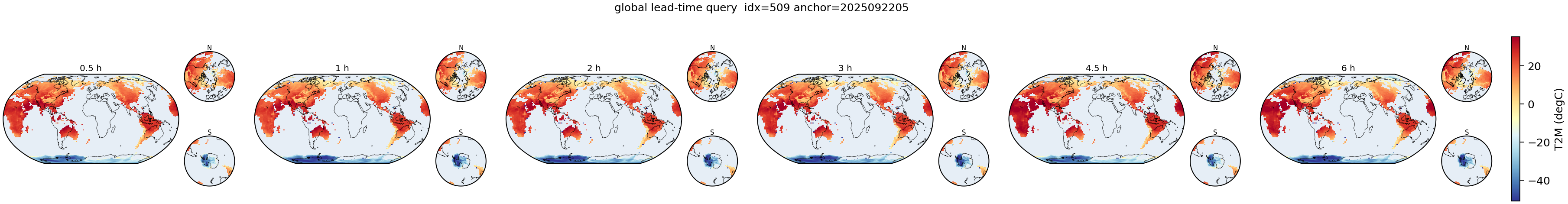}
  \caption{
  Continuous lead-time query diagnostics over the global domain ($180.0^\circ$W--$180.0^\circ$E, $90.0^\circ$S--$90.0^\circ$N).
  The three rows are independent test initializations at 2025-09-05 05:00 UTC, 2025-09-13 17:00 UTC, and 2025-09-22 05:00 UTC, rendered with the same globe-style projection as the main-text global diagnostic.
  Within each row, the queried lead time is varied while the input state is fixed, allowing visual inspection of large-scale thermal evolution under the continuous-time interface.
  Non-integer leads are forecast-only diagnostics because the current target split provides hourly references.
  }
  \label{fig:app_lead_query_global_cases}
\end{figure*}

\begin{figure*}[t]
  \centering
  \includegraphics[width=0.98\textwidth]{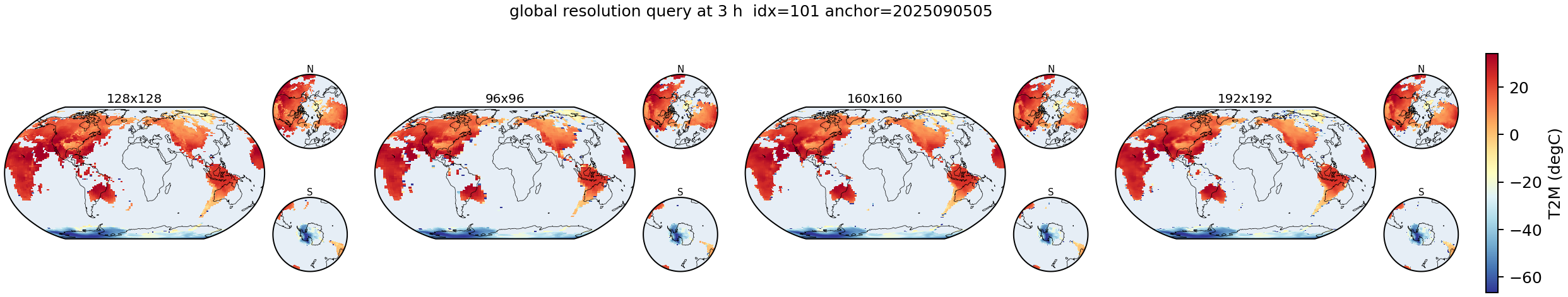}\\[2mm]
  \includegraphics[width=0.98\textwidth]{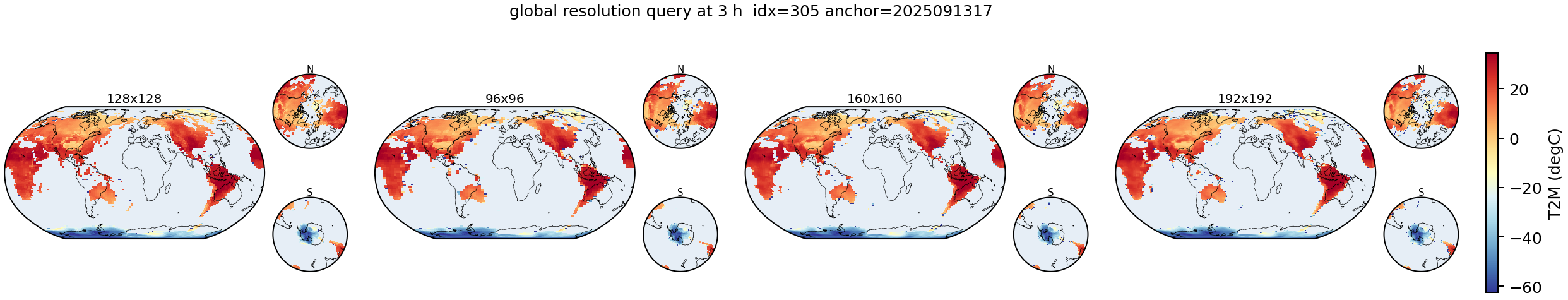}\\[2mm]
  \includegraphics[width=0.98\textwidth]{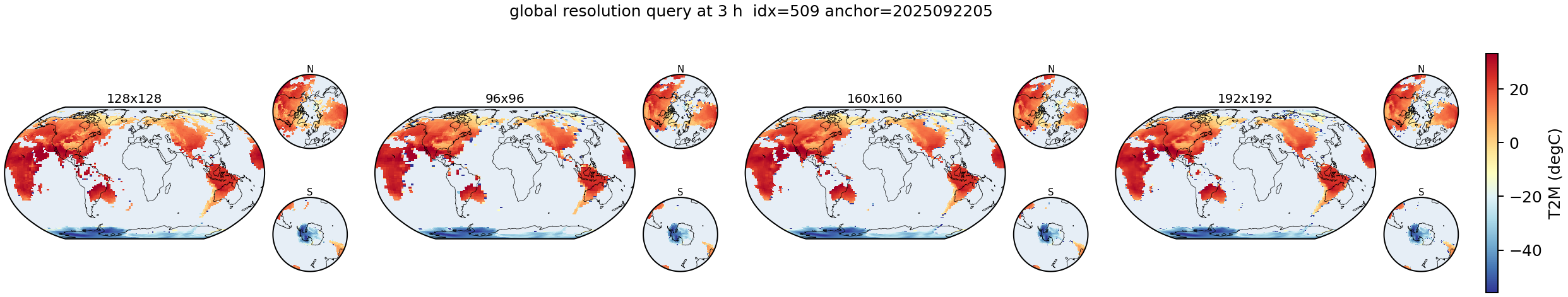}
  \caption{
  Resolution-conditioned query diagnostics over the global domain ($180.0^\circ$W--$180.0^\circ$E, $90.0^\circ$S--$90.0^\circ$N).
  The three rows are independent test initializations at 2025-09-05 05:00 UTC, 2025-09-13 17:00 UTC, and 2025-09-22 05:00 UTC.
  Within each row, the model is queried at multiple output grid sizes for the same forecast case, so changes across columns reflect direct resolution-conditioned field evaluation.
  These visualizations are used as qualitative checks of large-scale coherence under resolution queries, whereas quantitative cross-resolution consistency is reported separately in the main tables.
  }
  \label{fig:app_resolution_query_global_cases}
\end{figure*}

\begin{figure*}[t]
  \centering
  \begin{minipage}{0.48\textwidth}
    \centering
    \includegraphics[width=\linewidth]{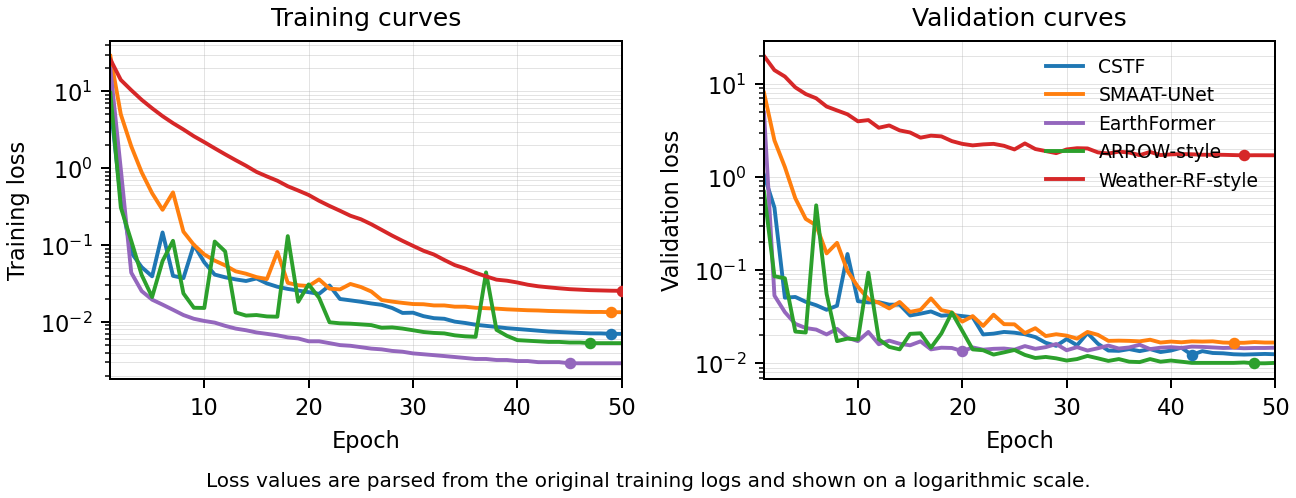}
  \end{minipage}
  \hfill
  \begin{minipage}{0.48\textwidth}
    \centering
    \includegraphics[width=\linewidth]{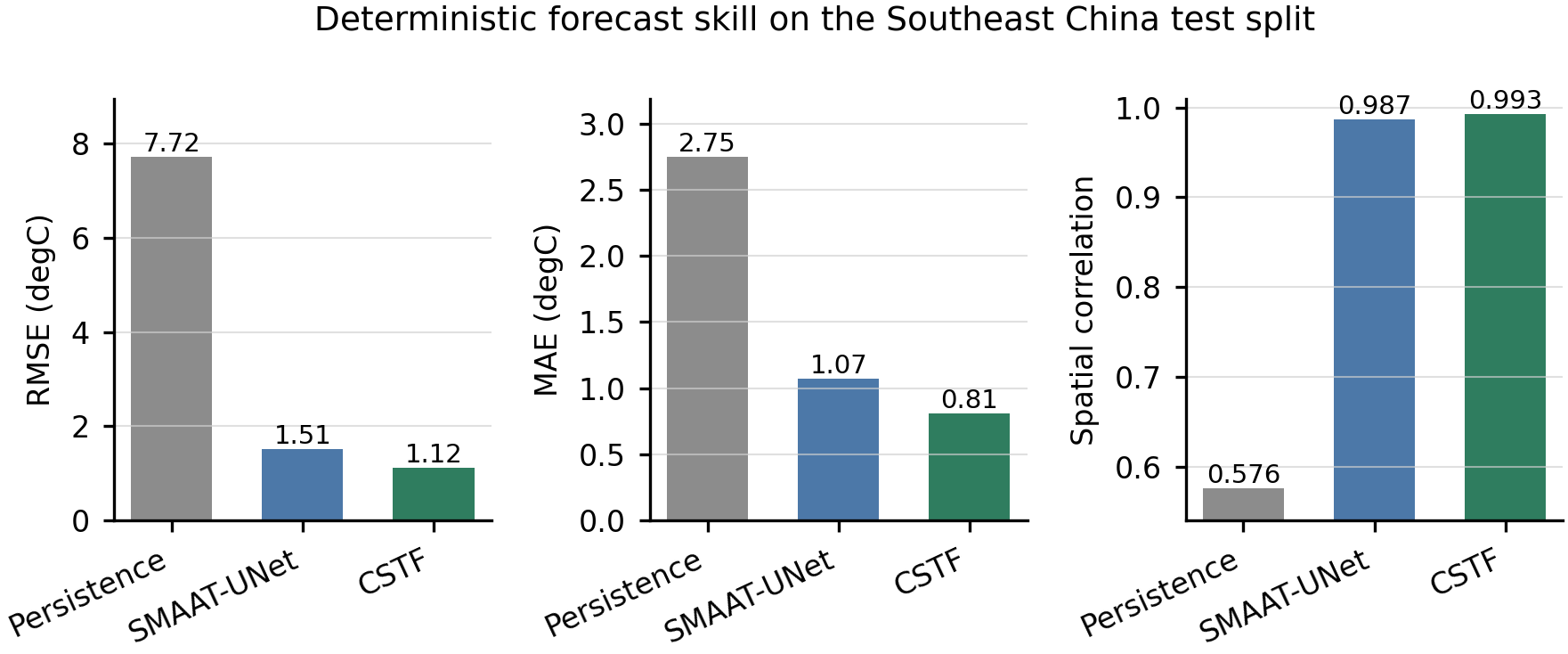}
  \end{minipage}\\[2mm]
  \begin{minipage}{0.48\textwidth}
    \centering
    \includegraphics[width=\linewidth]{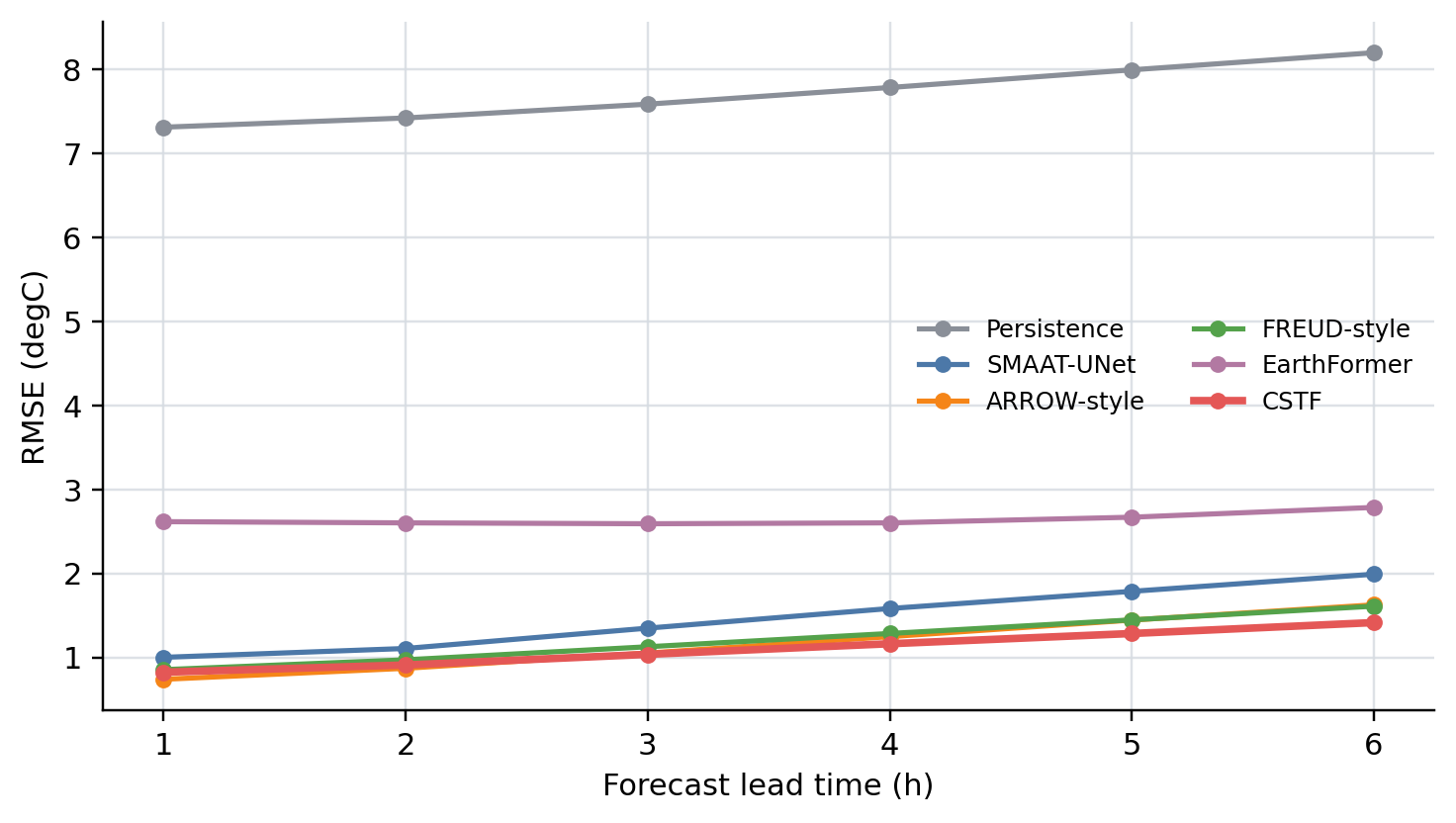}
  \end{minipage}
  \hfill
  \begin{minipage}{0.48\textwidth}
    \centering
    \includegraphics[width=\linewidth]{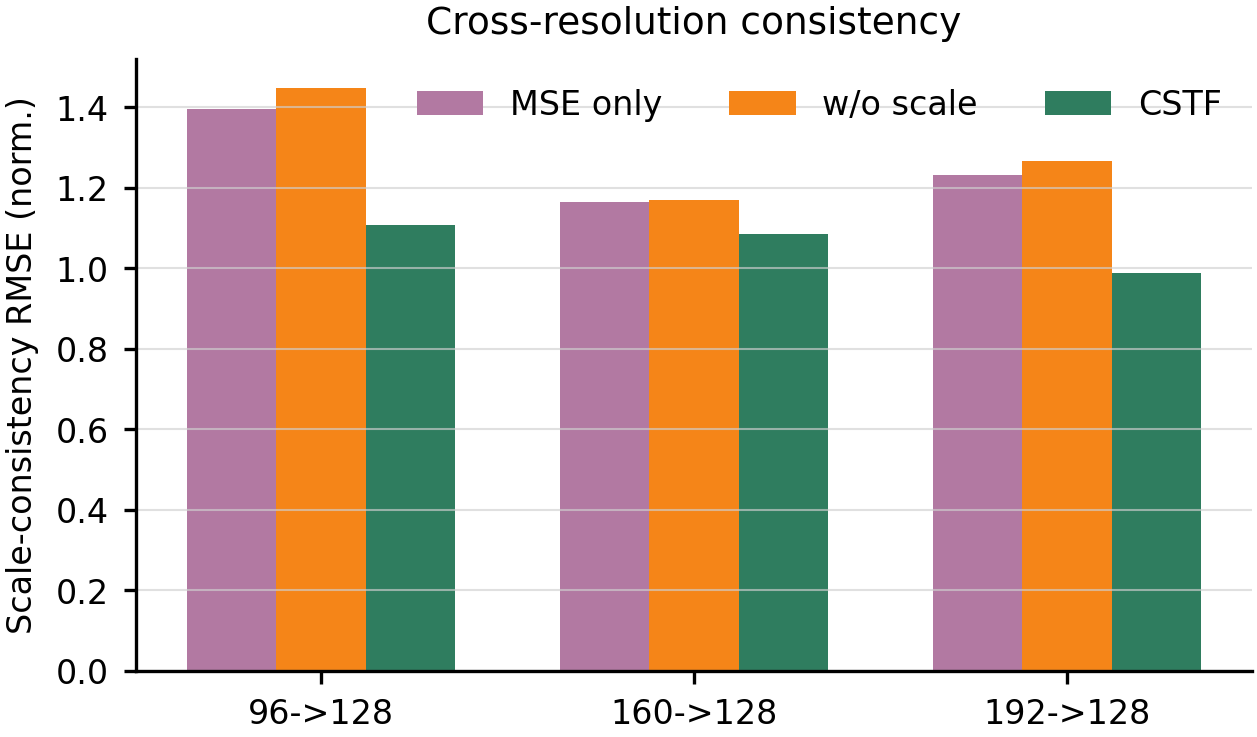}
  \end{minipage}\\[2mm]
  \begin{minipage}{0.48\textwidth}
    \centering
    \includegraphics[width=\linewidth]{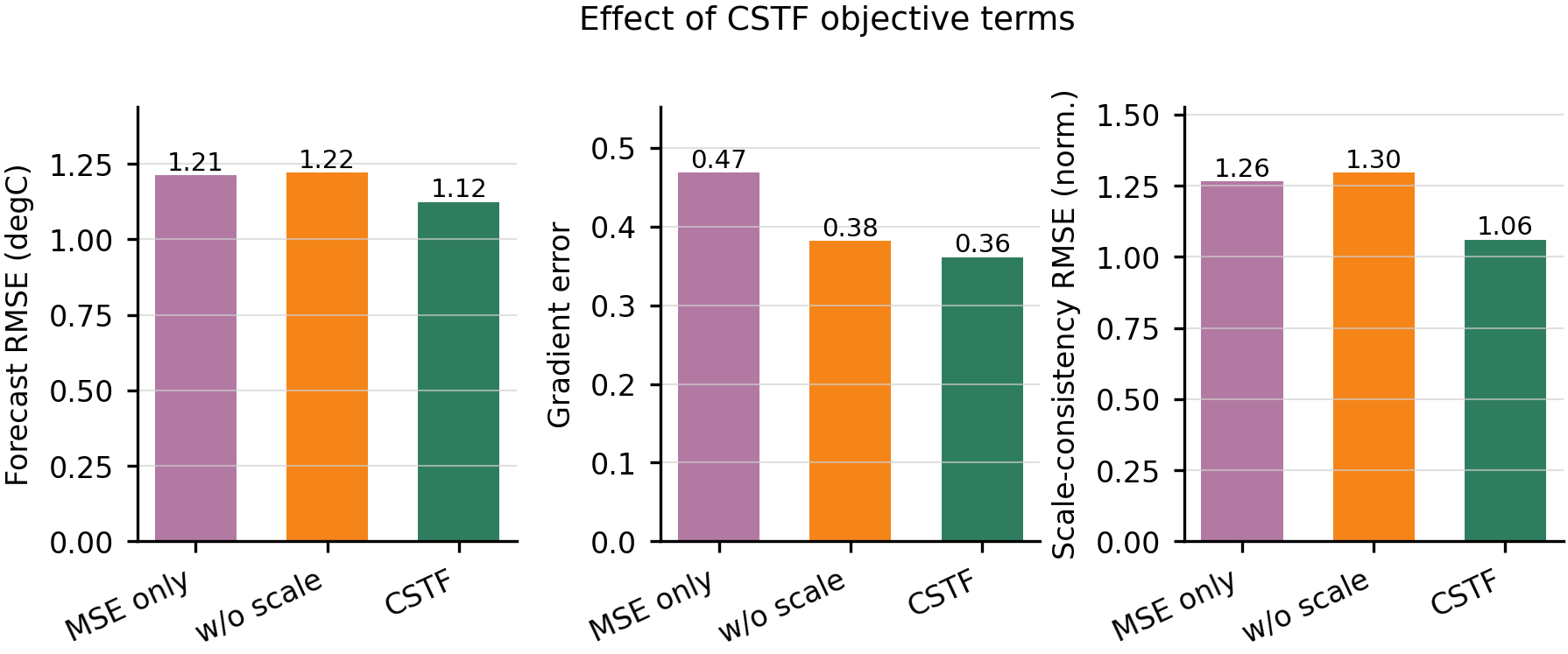}
  \end{minipage}
  \hfill
  \begin{minipage}{0.48\textwidth}
    \centering
    \includegraphics[width=\linewidth]{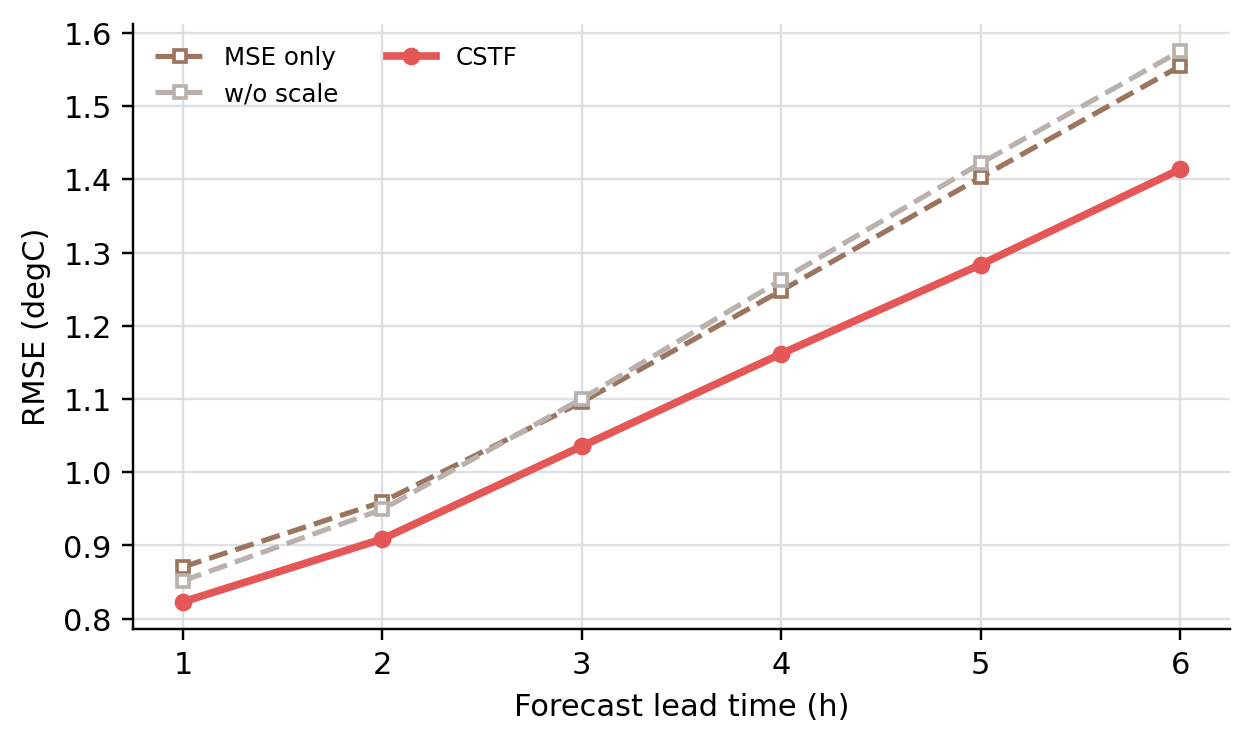}
  \end{minipage}
  \caption{
  Additional training, benchmark, and ablation diagnostics supporting the main quantitative results.
  From left to right and top to bottom, the panels show optimization curves, aggregate deterministic skill, lead-wise RMSE, cross-resolution scale consistency, objective-term ablation metrics, and lead-wise ablation behavior.
  The first panel checks whether model training is stable, the next two panels summarize forecast accuracy on the supervised hourly benchmark, the scale-consistency panel evaluates the resolution-query interface, and the ablation panels isolate the effects of the auxiliary objective terms.
  These plots are diagnostic summaries; the standardized test metrics and primary comparisons are reported in the main-text tables.
  }
  \label{fig:app_training_metric_diagnostics}
\end{figure*}

This section expands the evidence for the query-conditioned field interface beyond the main-text examples.
Figures~\ref{fig:app_lead_query_huadong_cases}--\ref{fig:app_training_metric_diagnostics} are organized to separate the two query modes and the two diagnostic domains.
Specifically, Figs.~\ref{fig:app_lead_query_huadong_cases} and \ref{fig:app_resolution_query_huadong_cases} examine continuous lead-time and resolution-conditioned behavior over Southeast China ($100.0^\circ$E--$120.0^\circ$E, $20.0^\circ$N--$40.0^\circ$N), whereas Figs.~\ref{fig:app_lead_query_global_cases} and \ref{fig:app_resolution_query_global_cases} provide the corresponding global-domain diagnostics ($180.0^\circ$W--$180.0^\circ$E, $90.0^\circ$S--$90.0^\circ$N).
Figure~\ref{fig:app_training_metric_diagnostics} then summarizes training, benchmark, scale-consistency, and ablation evidence to connect these qualitative diagnostics with the quantitative claims in the main text.

Figure~\ref{fig:app_lead_query_huadong_cases} extends the regional continuous-time analysis from a single main-text example to three independent Southeast China forecast initializations.
The purpose of this figure is to test whether varying the lead-time query produces physically plausible, spatially coherent temperature evolution rather than isolated artifacts at non-integer query times.

Figure~\ref{fig:app_resolution_query_huadong_cases} evaluates the regional resolution-query interface under the same Southeast China setting.
By holding the atmospheric input and forecast case fixed while changing only the requested output grid, the figure checks whether \method{} behaves as a direct field evaluator instead of a fixed-grid predictor followed by post-hoc resizing.

Figure~\ref{fig:app_lead_query_global_cases} transfers the continuous lead-time diagnostic to the global domain.
This figure is not intended as a separate global benchmark; rather, it examines whether the same query-conditioned model produces coherent large-scale thermal structures when lead time is varied under a global diagnostic view.

Figure~\ref{fig:app_resolution_query_global_cases} provides the complementary global-domain resolution-query diagnostic.
It is used to inspect whether large-scale temperature patterns remain stable across requested output grids, which supports the interpretation of resolution control as query-conditioned field evaluation rather than a visualization-only operation.

Figure~\ref{fig:app_training_metric_diagnostics} collects the quantitative diagnostics that support the preceding visual evidence.
The panels cover optimization behavior, deterministic forecast skill, lead-wise degradation, cross-resolution consistency, and objective-term ablations, providing a compact check that the queryable behavior is accompanied by stable training and controlled test performance.

\subsection{Extended Objective-Term Ablation}
\label{app:extended_ablation}

The main text reports the primary objective-term ablation while emphasizing the complete \method{} objective and the removal of the scale-consistency term.
To make the contribution of the remaining auxiliary regularizers explicit, Table~\ref{tab:app_extended_ablation_leadwise} adds two controlled variants that remove the spatial-gradient loss or the temporal-difference loss individually.
All variants use the same architecture, nine-variable \era{} input history, Southeast China test split, checkpoint-selection rule, and deterministic evaluation protocol.

The extended results show that removing either regularizer degrades lead-wise deterministic skill relative to the full objective.
Removing $\mathcal{L}_{\mathrm{grad}}$ gives the largest deterioration, especially at longer lead times, indicating that preserving regional thermal gradients is important for maintaining spatial temperature structure.
Removing $\mathcal{L}_{\mathrm{temp}}$ is less damaging than removing $\mathcal{L}_{\mathrm{grad}}$, but it still increases RMSE and MAE across all supervised leads, supporting the role of temporal-difference supervision in stabilizing short-range evolution.
Together with the scale-consistency ablation in the main text, these results indicate that the auxiliary terms provide complementary regularization for spatial structure, lead-wise evolution, and cross-resolution coherence.

\end{document}